\documentclass[11pt, letterpaper, logo, onecolumn, copyright]{main}

\usepackage[authoryear, sort&compress, round]{natbib}

\usepackage[inkscapeformat=png]{svg}

\usepackage[most, breakable, skins]{tcolorbox}

\tcbuselibrary{skins}
\usepackage{lipsum}
\usepackage{tabularx}
\usepackage{afterpage}
\usepackage{booktabs}
\usepackage{xurl} 
\usepackage{subcaption}
\usepackage{makecell}
\usepackage{multirow}
\usepackage{bm}
\usepackage{multicol}
\usepackage{array}
\usepackage{float}
\usepackage{listings, listings-rust}
\usepackage{fontawesome5}
\usepackage{hyperref}
\usepackage{amssymb}
\usepackage{graphicx}
\usepackage[dvipsnames]{xcolor}
\usepackage{cleveref}
\usepackage{longtable}
\usepackage{pdflscape}
\usepackage{adjustbox}
\usepackage{nicematrix}
\usepackage{CJKutf8}
\usepackage{ragged2e}
\usepackage{colortbl}
\usepackage{enumitem}
\usepackage{bbm}
\usepackage{pifont}
\usepackage{hyphenat}
\usepackage{amsmath}
\usepackage{amsthm}  
\usepackage{circledsteps}
\usepackage{diagbox}
\usepackage{bookmark}
\usepackage{textcomp}

\usepackage{algorithm} 
\usepackage{algpseudocode}
\providecommand{\STATE}{\State}
\providecommand{\IF}{\If}
\providecommand{\ENDIF}{\EndIf}

\usepackage{wrapfig}
\usepackage{xspace}
\usepackage{tikz}
\usepackage[normalem]{ulem}

\usepackage{pgfplots}
\usepackage{pgfplotstable}
\pgfplotsset{compat=1.18}

\usepackage[most]{tcolorbox}
\definecolor{lightblue}{RGB}{210, 220, 250}

\definecolor{medgray55}{gray}{0.55}
\definecolor{medgray}{gray}{0.7}
\definecolor{litegray}{gray}{0.9}
\definecolor{gblue}{RGB}{210, 227, 252}
\definecolor{gred}{RGB}{250, 210, 207}
\definecolor{gyellow}{RGB}{254, 239, 195}
\definecolor{ggreen}{RGB}{206, 234, 214}
\definecolor{gorange}{RGB}{254, 223, 200}

\definecolor{gblue9}{RGB}{23, 78, 166}
\definecolor{gred9}{RGB}{165, 14, 14}
\definecolor{gyellow9}{RGB}{227, 116, 0}
\definecolor{ggreen9}{RGB}{13, 101, 45}
\definecolor{gorange9}{RGB}{176, 96, 0}

\definecolor{myblue}{rgb}{0,0,1}
\definecolor{myred}{rgb}{1,0,0}
\definecolor{mylightgray}{gray}{0.95}
\definecolor{myCite}{HTML}{1C4587}

\definecolor{highlightblue}{HTML}{185ABC}
\definecolor{cellHighlight}{HTML}{dbefff}
\definecolor{lightgray}{RGB}{211, 211, 211}
\definecolor{lightfont}{gray}{0.3}

\usepackage{minitoc}

\noptcrule


\newcolumntype{L}[1]{>{\raggedright\let\newline\\\arraybackslash\hspace{0pt}}m{#1}}
\newcolumntype{C}[1]{>{\centering}m{#1}}

\newcolumntype{R}[1]{>{\raggedleft\let\newline\\\arraybackslash\hspace{0pt}}m{#1}}

\definecolor{ao}{rgb}{0.0, 0.0, 1.0}

\newcommand\vcent[1]{\vcenter{\hbox{#1}}}

\newcommand\loudspeaker[1][3]{\ensuremath{\vcent{\rule{.6ex}{.6ex}}\kern-.5ex
  \vcent{\scalebox{.6}[1]{\rotatebox[origin=center]{90}{$\blacktriangle$}}}
  \ifnum#1>0\relax\kern.05ex\vcent{\scalebox{.4}{\ttfamily)}}
  \ifnum#1>1\relax\kern-.4ex\vcent{\scalebox{.56}{\ttfamily)}}
  \ifnum#1>2\relax\kern-.55ex\vcent{\scalebox{.7}{\ttfamily)}}
  \fi\fi\fi}
}

\makeatletter
\renewcommand\subparagraph{
 \@startsection {subparagraph}{5}{\z@ }{3.25ex \@plus 1ex
 \@minus .2ex}{-1em}{\normalfont \normalsize \bfseries }}
\makeatother

\let\cite\citep
\hypersetup{
  citecolor = myCite,  
  linkcolor = myCite,   
  urlcolor  = myCite
}
\newcommand{\myheaderbreak}{\\}
\title{LLMEval-Fair: A Large-Scale Longitudinal Study on Robust\myheaderbreak and Fair Evaluation of Large Language Models}

\author{
  Ming Zhang\textsuperscript{\rm *\dag},
  Yujiong Shen\textsuperscript{\rm *},
  Jingyi Deng\textsuperscript{\rm *},
  Yuhui Wang\textsuperscript{\rm *},
  Huayu Sha\textsuperscript{\rm },\\
  Kexin Tan\textsuperscript{\rm },
  Qiyuan Peng\textsuperscript{\rm },
  Yue Zhang\textsuperscript{\rm },
  Junzhe Wang\textsuperscript{\rm },
  Shichun Liu\textsuperscript{\rm },
  Yueyuan Huang\textsuperscript{\rm },\\
  Jingqi Tong\textsuperscript{\rm },
  Changhao Jiang\textsuperscript{\rm },
  Yilong Wu\textsuperscript{\rm },
  Zhihao Zhang\textsuperscript{\rm },
  Mingqi Wu\textsuperscript{\rm },
  Mingxu Chai\textsuperscript{\rm },\\
  Zhiheng Xi\textsuperscript{\rm },
  Shihan Dou\textsuperscript{\rm },
  Tao Gui\textsuperscript{\rm },
  Qi Zhang\textsuperscript{\rm \dag},
  Xuanjing Huang\textsuperscript{\rm }\\
  \vspace{0.3cm} 
  \normalsize 
  \textsuperscript{\rm}Fudan NLP Group \\ 
  \texttt{\normalsize mingzhang23@m.fudan.edu.cn, qz@fudan.edu.cn}
}

\vspace{-50pt}
\begin{abstract}
Existing evaluation of Large Language Models (LLMs) on static benchmarks is vulnerable to data contamination and leaderboard overfitting, critical issues that obscure true model capabilities. To address this, we introduce LLMEval-Fair, a framework for dynamic evaluation of LLMs. LLMEval-Fair is built on a proprietary bank of 220k graduate-level questions, from which it dynamically samples unseen test sets for each evaluation run. Its automated pipeline ensures integrity via contamination-resistant data curation, a novel anti-cheating architecture, and a calibrated LLM-as-a-judge process achieving 90\% agreement with human experts, complemented by a relative ranking system for fair comparison. A 30-month longitudinal study of nearly 60 leading models reveals a performance ceiling on knowledge memorization and exposes data contamination vulnerabilities undetectable by static benchmarks. The framework demonstrates exceptional robustness in ranking stability and consistency, providing strong empirical validation for the dynamic evaluation paradigm. LLMEval-Fair offers a robust and credible methodology for assessing the true capabilities of LLMs beyond leaderboard scores, promoting the development of more trustworthy evaluation standards.Our code and data are publicly available at \url{https://github.com/llmeval/LLMEval-Fair}.
\end{abstract}

\begin{document}
\doparttoc
\faketableofcontents

\begingroup
  \renewcommand\thefootnote{}
  \footnote{\hspace{-1.8em}\textsuperscript{*}Equal Contribution.\\
            \textsuperscript{\dag}Corresponding Authors.}
\endgroup

\vspace{-30pt}
\maketitle
\renewcommand{\myheaderbreak}{ }

\vspace{-15pt}
\section{Introduction}
The rapid advancement of Large Language Models (LLMs) has led to a proliferation of benchmarks designed to assess their capabilities \cite{DBLP:journals/corr/abs-2307-03109,DBLP:conf/emnlp/LaskarABRKKJB0P24,DBLP:journals/corr/abs-2506-11094}. However, these benchmarks predominantly rely on a static evaluation paradigm where models are tested on fixed, publicly available datasets \cite{DBLP:journals/corr/abs-2503-04149}. This approach is fundamentally vulnerable to data contamination and test set overfitting, contributing to a growing ``evaluation crisis'' where benchmark scores may no longer reliably reflect a model's generalizable abilities \cite{DBLP:journals/corr/abs-2412-03597,DBLP:conf/naacl/Deng0TGC24,DBLP:journals/corr/abs-2402-02823}. For example, GPT-4 achieved exact match rates of 52\% and 57\% when guessing the masked options in MMLU \cite{mmlu} test sets, far exceeding random chance \cite{leakage:deng2024}. Similarly, Qwen-1.8B has been shown to exactly replicate complete n-grams from both training and test splits of GSM8K \cite{GSM8K} and MATH \cite{MATH}, including 25 exact 5-gram matches in the MATH test set, indicating potential undetected data leakage and memorization \cite{benchmarkleakage:xu2024}. Furthermore, static benchmark designs, dataset contamination, and biased evaluation protocols could create misleading perceptions of LLM capabilities, undermining the reliability of current performance assessments \cite{DBLP:journals/corr/abs-2412-03597}.

The crisis compels us to shift our focus from what capabilities to evaluate, such as knowledge and reasoning, to a more foundational question: how to evaluate in a manner that is robust, fair, and resistant to strategic manipulation. Based on our analysis of the current evaluation crisis, we identify three fundamental challenges for constructing a trustworthy evaluation framework. These challenges correspond to the core stages of any assessment: the data, the protocol, and the ranking system.

\begin{figure*}[t]   
    \centering     
    \includegraphics[width=1.00\textwidth]{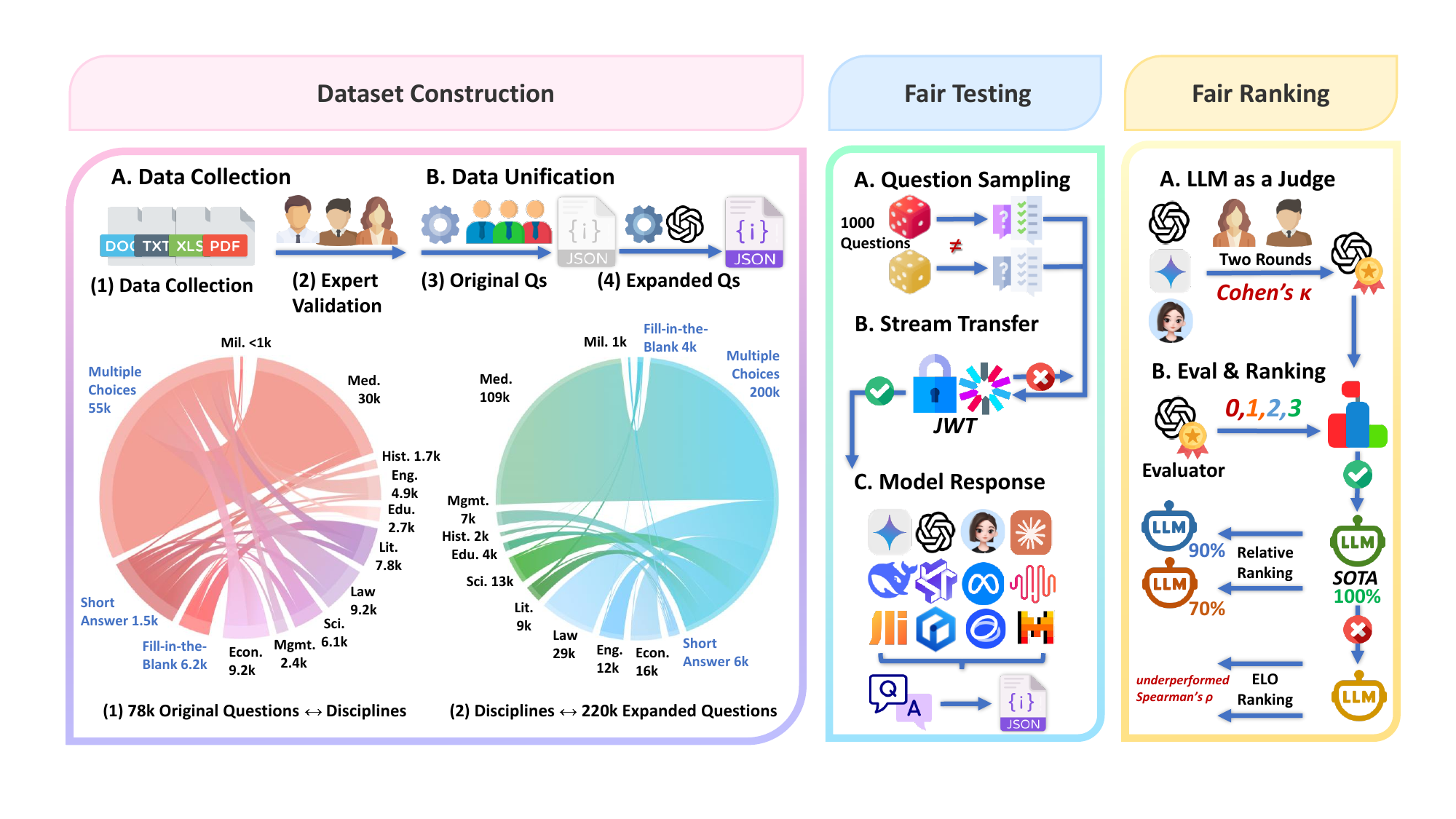}
    \caption{The LLMEval-Fair framework comprises three core stages. First, in Data Construction, diverse exam data are collected, filtered by experts, standardized into original questions in JSON format, and expanded to enhance formality and diversity. Second, Fair Testing involves sampling 1,000 unique questions for models’ evaluation, transmitting data via a secure JWT stream, and collecting model responses. Third, Fair Ranking begins by selecting evaluators based on human-machine agreement, measured by Cohen's $\kappa$ coefficients. Subsequently, the process entails generating relative rankings, validating robustness through ablation studies, and analyzing model errors. Discipline abbreviations: Eng. (Engineering), Econ. (Economics), Edu. (Education), Lit. (Literature), Mgmt. (Management), Sci. (Science), Hist. (History), Med. (Medicine), Mil. (Military).}
\label{fig:main-graph}
\end{figure*}

\textbf{Challenge 1: How can we ensure the integrity of the evaluation data?} The cornerstone of any benchmark is its test set. If the data are public or predictable, they become susceptible to leakage into model training corpora, leading to inflated scores that reflect memorization rather than true capability. Therefore, building a benchmark that is inherently resilient to data contamination is the primary challenge. To address this, we construct a private question bank of over 220k graduate-level questions, augmented with structural variations to mitigate memorization.

\textbf{Challenge 2: How can we design an unpredictable assessment protocol?} Even with private data, a static protocol with fixed test sets can be reverse-engineered or exploited over time. A truly robust evaluation requires a dynamic and unpredictable process that prevents strategic games. 
To achieve this, we implement a dynamic evaluation protocol where models are served unseen, randomly sampled questions in each session through a secure, two-layer anti-cheating architecture, ensuring that each evaluation is unique and unpredictable.

\textbf{Challenge 3: How can we establish a fair and stable ranking system under dynamic conditions?} When the evaluation content is not fixed, traditional absolute scoring becomes unreliable for cross-model comparison. A fair ranking system must remain stable and consistent, even when models are tested on different, albeit equivalent, sets of questions. 
To solve this, we develop a novel relative ranking system powered by a highly-calibrated LLM-as-a-Judge framework. This system ranks models by comparing their performance within each evaluation session rather than relying on absolute scores, ensuring fair rankings with negligible variance across different data samples.

Collectively, these solutions constitute LLMEval-Fair, a comprehensive framework for dynamic LLM evaluation. We leverage this framework to conduct an extensive longitudinal study on our official platform, spanning from the first half of 2023 to the second half of 2025. The study is longitudinal across three simultaneous dimensions: \textit{temporal} (continuous tracking over 30 months rather than single snapshots), \textit{data} (the question bank is incrementally expanded, so later evaluation rounds sample from a substantively different pool than earlier ones), and \textit{model} (successive generations within the same family, e.g., GPT-3.5 $\to$ GPT-4o $\to$ GPT-5, are compared under a unified protocol). Throughout this period, we have continuously evaluated nearly 60 leading proprietary and open-source models, typically assessing them within one week of their public release. Each model is pinned to a specific, fixed API snapshot (e.g., \texttt{gpt-4o-2024-11-20}) and undergoes at least three independent, randomly sampled evaluation runs to ensure stability. To date, this rigorous, ongoing effort has accumulated over 180k evaluation data points.

Leveraging this large-scale evaluation data, we systematically investigate three research questions corresponding to our core challenges and find that (a) all models converge to a performance ceiling around 90\% persistent gaps in specialized domains like literature and medicine, (b) dynamic rankings diverge from static benchmarks, and static benchmarks suffer from severe data contamination, (c) our framework demonstrates exceptional ranking stability with negligible variance under multi-round resampling and varying sample sizes. More insightful findings are presented in Section \ref{Experiment and Analysis}.

Overall, our contributions are threefold:
\begin{enumerate}
    \item We construct LLMEval-Fair, a large-scale anti-cheating evaluation platform featuring a proprietary 220k question bank, secure dynamic sampling protocols, and robust anti-manipulation mechanisms for trustworthy LLM assessment.
    \item We conduct an extensive 30-month longitudinal evaluation campaign across nearly 60 leading models, accumulating over 180k evaluation data points through continuous anti-cheating assessments and comparative analysis with static benchmarks.
    \item We conduct extensive empirical analysis across three research questions corresponding to our core challenges, revealing eight key findings about model performance ceilings, ranking stability, and contamination vulnerabilities in current evaluation practices.
\end{enumerate}

\section{Design}

In this section, we detail the design and implementation of LLMEval-Fair, which addresses the three fundamental challenges through a three-stage framework. \textit{Dataset Construction} tackles data integrity by building a contamination-resistant private question bank. \textit{Evaluation Process} addresses unpredictable assessment through dynamic sampling and anti-cheating mechanisms. \textit{Ranking System} implements a calibrated ranking system for fair model comparison.

\subsection{Dataset Construction}
To build a contamination-resistant and high-quality question bank, we sourced postgraduate and undergraduate exam questions from Chinese universities, covering 13 primary and over 50 secondary academic disciplines. Figure~\ref{fig:main-graph} offers a comprehensive overview of the design, highlighting three key stages: Dataset Construction, Fair Testing and Fair Ranking.

All questions and evaluation prompts are in Chinese, as the benchmark targets Chinese-language capabilities; translated examples are provided in the appendix for international readers. The construction follows a rigorous pipeline. First, we collect original exam questions from diverse formats and invite over 30 graduate-student annotators with domain expertise corresponding to the question disciplines for quality screening. Each item is independently reviewed by two annotators before finalization. This dual-review process yields 78,009 high-quality original questions after eliminating those with factual errors or irrelevant answers. Second, we employ an LLM-driven augmentation process to expand coverage and diversity. For instance, each Multiple-Choice question with $n$ options is converted into $n$ Fill-in-the-Blank variants, while Material Analysis questions are decomposed into multiple true/false questions based on key information. Finally, all augmented questions undergo format verification and metadata enrichment to ensure quality and traceability. A controlled comparison confirms that augmentation preserves evaluation validity: model rankings remain identical (Spearman $\rho = 1.0$) with no statistically significant score difference ($p = 0.736$, Cohen's $d = 0.16$); details are in Appendix~\ref{sec:augmentation_validation}.

As of early 2025, this process has resulted in the \textbf{LLMEval-Fair dataset}, which comprises over 220k questions across six main categories. The full dataset covers all 13 primary disciplines; for evaluation, we select the 10 most data-sufficient disciplines to ensure statistically balanced cross-discipline comparisons, excluding Agronomy, Arts, and Philosophy due to insufficient question volumes at the time of initial sampling.\footnote{Philosophy questions were later collected in sufficient quantity but were not introduced mid-evaluation to maintain cross-model fairness.} To maintain evaluation freshness and prevent contamination, we continuously expand the question bank through the same manual collection and automated augmentation pipeline. Detailed statistical analysis of the dataset, including disciplinary breakdown and content diversity, is provided in Appendix~\ref{sec:Dataset}.

\subsection{Evaluation Process}

To ensure a reliable and fair evaluation, we design a dynamic process centered on a multi-layered anti-cheating architecture. This approach guarantees that each evaluation is unique, robust against manipulation, and accurately reflects a model's capabilities. The process is built upon two core strategies: dynamic question sampling and a secure delivery architecture.

\subsubsection{Dynamic Question Sampling}
To ensure unpredictability, each model evaluation is based on a unique set of 1,000 questions sampled from our private question bank via a three-stage stratified procedure. \textbf{Stage~1 (Discipline-level quota):} We allocate approximately 100 questions to each of the 10 evaluated disciplines, ensuring balanced coverage. \textbf{Stage~2 (Question-type allocation):} Within each discipline's quota, questions are allocated proportionally by type (e.g., multiple-choice, short-answer) according to the type distribution in the augmented dataset. \textbf{Stage~3 (Sub-discipline sampling):} Questions are randomly drawn from each secondary discipline in proportion to its share within the discipline--type stratum; both original and augmented questions are mixed in the sampling pool. After sampling, each question enters a three-state lifecycle (\textit{allocated} $\to$ \textit{pending} $\to$ \textit{completed}) managed by the inner-layer process control (Section~2.2.2), preventing re-use within or across sessions. Models must answer questions in the pre-allocated order, preventing any ``cherry-picking'' strategies. This ensures that every evaluation is a distinct event reflecting the model’s true generalization ability.

\subsubsection{Secure Anti-Cheating Architecture}
The evaluation process is protected by a secure, two-layer anti-cheating architecture. 
\begin{itemize}
    \item \textbf{The Outer Layer (Access Control):} This layer manages authentication and authorization. We use JSON Web Tokens (JWT)\cite{rfc7519} to secure every API request, ensuring that only authenticated models can participate in an evaluation session. A strict Role-Based Access Control (RBAC) system prevents any cross-session or cross-user data access, isolating each evaluation.
    \item \textbf{The Inner Layer (Process Control):} This layer enforces the evaluation rules. A multi-level quota system tracks the number of questions allocated, pending, and completed, effectively preventing models from attempting to acquire more questions than permitted or resubmitting answers. As a final safeguard, our system automatically strips all answers and explanations from the data transmitted to the model, ensuring that only the question content is exposed and preventing answer leakage through data parsing.
\end{itemize}

\subsection{Ranking System}

To establish fair and stable rankings under dynamic evaluation conditions, we develop a calibrated ranking framework that combines LLM-as-a-Judge evaluation with relative scoring mechanisms. This approach ensures consistent and reliable model comparisons even when different question sets are used across evaluation sessions.

\subsubsection{LLM-as-a-Judge Evaluation}

To quantify answer quality, we establish a standardized scoring metric with an integer range of [0,3], ensuring consistent evaluation of response efficacy across diverse model architectures. For scoring implementation, we uniformly employ GPT-4o \cite{OpenAI2023GPT4} as our judge, which has demonstrated high human-machine agreement through rigorous validation (detailed in Section \ref{Research Question III}). This choice of a single, validated scoring model eliminates potential biases introduced by varying evaluative criteria.

The scoring focuses on both core correctness and explanation quality, with core correctness serving as the primary indicator for score determination. The specific evaluation prompt and criteria are provided in Appendix~\ref{sec:prompts}.

\subsubsection{Evaluation Metrics}

To mitigate systematic bias introduced by random sampling questions, LLMEval-Fair employs both relative score and absolute scores as evaluation metrics.

The absolute score $S_{\text{model}}$ represents a model's performance on $N=1000$ questions, where each question receives a score $s_i$ (with maximum score $s_{\text{max}}=3$), mapped to the [0,100] interval:

\begin{equation}
S_{\text{model}} = \frac{\sum_{i=1}^N s_i}{N \times s_{\text{max}}} \times 100
\end{equation}

The relative score $R_{\text{SOTA}}^{\text{model}}$ is defined as the model's absolute score relative to the current SOTA model's absolute score on the same question set, mapped to the [0,100] interval:

\begin{equation}
R_{\text{SOTA}}^{\text{model}} = \frac{S_{\text{model}}}{S_{\text{SOTA}}} \times 100
\end{equation}

In our current evaluation, we use \textit{Doubao-1.5-Thinking-Pro} as the reference SOTA model, as it achieved the highest absolute score (93.67) among all evaluated models within our unified protocol while exhibiting the highest multi-round stability (variance $= 0.00$ across three independent 1,000-question trials; see Appendix~\ref{sec:detailsInExperiment}). This is an operational designation: should a stronger model emerge, the anchor can be updated and all relative scores recalculated, as the formula is anchor-agnostic by design. 

\begin{table*}[t]
  \centering  
  \scalebox{0.80}{%
  \begin{tabular}{lcccccccccccc}
    \hline
    Model & $R_{\text{SOTA}}^{\text{model}}$ & $S_{\text{model}}$ & Eng. & Econ. & Edu. & Law & Lit. & Mgmt. & Sci. & Hist. & Med. & Mil. \\
    \hline
    \multicolumn{13}{l}{\textit{Open-source LLMs}} \\
    DeepSeek-R1                   & 97.40 & \textbf{91.23} & \textbf{9.47} & 9.43 & \textbf{9.27} & 9.37 & \textbf{8.83} & 9.37 & \textbf{9.03} & \textbf{9.53} & 8.50 & 8.43 \\
    DeepSeek-V3                   & 96.47 & 90.36 & 9.30 & \textbf{9.57} & 8.93 & 9.23 & 8.60 & 9.13 & 8.97 & 9.47 & \textbf{8.83} & 8.33 \\
    Qwen-3-235B                  & 96.42 & 90.32 & 9.23 & 9.43 & 9.03 & \textbf{9.50} & 8.23 & 9.43 & 8.97 & 9.17 & 8.73 & \textbf{8.60} \\
    Qwen-3-32B                   & 92.22 & 86.38 & 8.43 & 9.10 & 8.57 & 9.10 & 7.77 & \textbf{9.47} & 8.67 & 9.30 & 7.70 & 8.27 \\
    \hline
    \multicolumn{13}{l}{\textit{Closed-source LLMs}} \\
    Doubao-1.5-Thinking-Pro      & \textbf{100.00} & \textbf{93.67} & \textbf{9.47} & \textbf{9.67} & \textbf{9.43} & \textbf{9.77} & \textbf{8.93} & 9.53 & \textbf{9.23} & \textbf{9.70} & \textbf{8.97} & \textbf{8.97} \\
    Gemini-2.5-Pro               & 97.22 & 91.07 & 9.20 & 9.47 & 9.20 & 9.30 & 8.43 & \textbf{9.63} & 9.07 & 9.40 & 8.50 & 8.87 \\
    Gemini-2.5-Pro-Thinking      & 97.15 & 91.00 & 9.13 & 9.50 & 9.37 & 9.47 & 8.40 & \textbf{9.63} & 9.20 & 9.27 & 8.30 & 8.73 \\
    Doubao-1.5-Pro               & 95.68 & 89.62 & 8.83 & 9.03 & 9.13 & 9.43 & 8.57 & 9.27 & 8.83 & 9.10 & 8.60 & 8.83 \\
    Kimi-K2                        & 94.27 & 88.30 & 9.23 & 9.17 & 8.80 & 9.00 & 8.40 & 9.17 & 8.77 & 9.13 & 8.53 & 8.10 \\
    GPT-5                          & 93.84 & 87.90 & 8.83 & 9.37 & 8.90 & 8.87 & 8.10 & 9.10 & 8.90 & 9.03 & 8.50 & 8.30 \\
    Claude-Sonnet-4.5-Thinking     & 93.48 & 87.57 & 8.90 & 9.17 & 8.80 & 8.97 & 8.00 & 9.23 & 8.90 & 9.00 & 8.27 & 8.33 \\
    o1                       & 93.36 & 87.45 & 8.90 & 9.30 & 8.67 & 8.77 & 7.73 & 9.27 & 8.90 & 8.97 & 8.17 & 8.77 \\
    Claude-Sonnet-4-Thinking       & 91.03 & 85.27 & 8.57 & 9.00 & 8.63 & 8.73 & 7.57 & 9.10 & 8.93 & 8.70 & 7.97 & 8.07 \\
    Claude-Sonnet-4                & 91.00 & 85.24 & 8.57 & 8.80 & 8.50 & 8.70 & 7.80 & 9.03 & 8.80 & 8.80 & 8.17 & 8.07 \\
    GPT-4o-search                & 89.40 & 83.74 & 8.27 & 8.77 & 8.43 & 8.67 & 7.77 & 8.80 & 8.20 & 8.73 & 8.27 & 7.83 \\
    GPT-4o                       & 88.09 & 82.51 & 7.90 & 8.67 & 8.30 & 8.33 & 7.17 & 8.97 & 8.57 & 8.67 & 7.63 & 8.30 \\
    o3-mini                        & 84.13 & 78.80 & 7.97 & 8.60 & 8.30 & 8.20 & 6.73 & 8.57 & 8.53 & 7.17 & 7.03 & 7.70 \\

    \hline
  \end{tabular}
  }
  \caption{Overall and Subject‐Level Scores. $R_{\text{SOTA}}^{\text{model}}$ represents the relative score (0-100 scale) as defined in Equation (2), with Doubao-1.5-Thinking-Pro as the reference SOTA model. $S_{\text{model}}$ represents the absolute score (0-100 scale) as defined in Equation (1). Subject-level scores use a 10-point scale. }
\label{tab:overall-ranking}
\end{table*}

\section{Experiment and Analysis}
\label{Experiment and Analysis}

Based on the three core challenges identified in our introduction, we design the LLMEval-Fair evaluation framework. In this section, we systematically investigate three critical research questions that arise from these challenges:

\textbf{Research Question I:} What authentic capability distributions and longitudinal trends do LLMs exhibit under LLMEval-Fair?

\textbf{Research Question II:} How does LLMEval-Fair dynamic evaluation compare with static benchmarks regarding ranking accuracy and contamination issues?

\textbf{Research Question III:} How stable and reliable is LLMEval-Fair's relative ranking system under multi-round resampling and human-machine consistency validation?

Through comprehensive experiments designed to address these research questions, we aim to validate the effectiveness of our dynamic evaluation paradigm and provide empirical evidence for the superiority of contamination-resistant assessment frameworks.

\subsection{Experimental Setup}
\paragraph{Benchmarking LLMs on LLMEval-Fair}  
We tracked nearly 60 LLMs from June 2023 to December 2025, presenting full results in Appendix~\ref{sec:Leaderboard} and focusing here on 17 representative models (proprietary and open-source). Appendix~\ref{app:model_selection} provides a detailed overview of these representative models. Each model is evaluated across three prompting paradigms (Zero-Shot, Few-Shot, Chain-of-Thought) and 10 academic disciplines. In addition, we sample incorrect responses from all evaluated models and manually classify failures into five categories: disciplinary knowledge, misunderstanding, logical reasoning, factual inaccuracies, and format compliance.

\paragraph{Ablation Studies}  
We conduct comprehensive ablation studies across two key dimensions:

\textbf{Benchmark Comparison:} We measure Spearman correlation between LLMEval-Fair and static benchmarks (AGIEval \cite{AGIEval}, C-Eval \cite{c-eval}) and perform fill-in-the-blank replay tests (1,000 questions, three attempts each) to assess contamination.  

\textbf{Ranking Validation:} We conduct multi-round resampling (n=1000, 2000, 4000) to test stability. Human-machine agreement validation involves two independent rounds of human evaluation with Cohen's $\kappa$ coefficients computed against three LLM-as-Judge evaluators. Second, we run an ablation study comparing our relative ranking to the traditional Elo scoring system. The introduction to the Elo scoring system can be found in Appendix~\ref{sec:elo}.

\subsection{Research Question I}

\begin{figure}[t]
\centering
\includegraphics[width=0.98\linewidth]{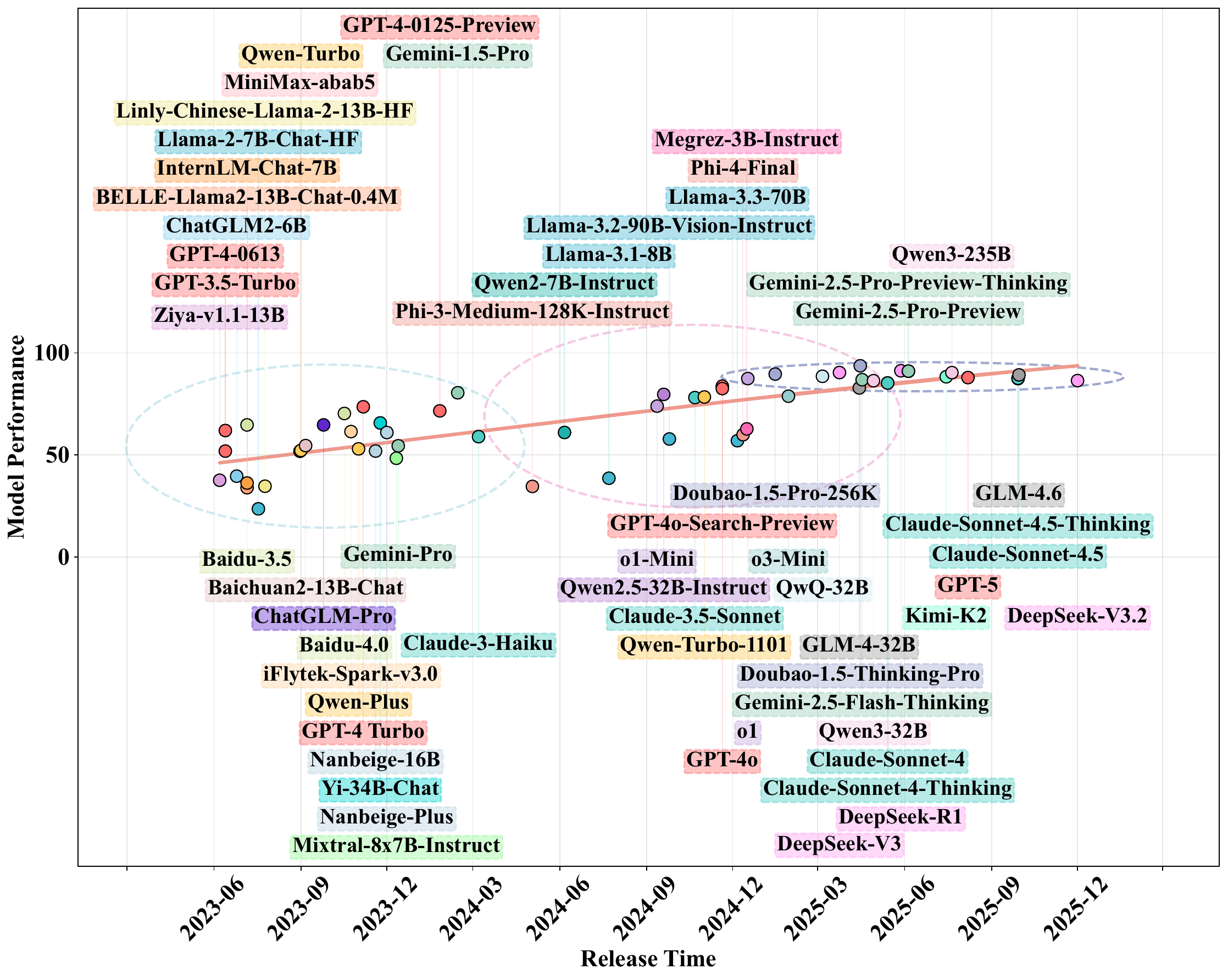}
\caption{Trend of model series. Models of the same series are primarily illustrated in the same color for better distinction. The fitted curve highlights the overall growth of performance over the observed period.}
\label{fig:Trend of model series.}
\end{figure}

\paragraph{Finding 1: All models converge to a performance ceiling of around 90\% over a longitudinal period, with leading open-source LLMs rivaling proprietary SOTA.} Figure~\ref{fig:Trend of model series.} illustrates the performance growth trajectories of different model series over time. Table~\ref{tab:overall-ranking} presents comprehensive evaluation scores and subject-level breakdowns for each model under the LLMEval-Fair framework. Nearly all models approach a performance ceiling around 90 on academic knowledge tasks. This convergence indicates fundamental limits in current model architectures for knowledge-intensive evaluation, consistent with theoretical analyses of knowledge retention upper bounds in pre-training \cite{Jiang2025BeyondScaling}.

The top proprietary model (Doubao-1.5-Thinking-Pro, 93.67) and the top open-source model (DeepSeek-R1, 91.23) both substantially outperform established systems such as GPT-4o (82.51), demonstrating that open-source LLMs can rival proprietary SOTA. Performance within a model family is not always monotonically increasing (e.g., DeepSeek-V3.2 scores below DeepSeek-V3); we discuss these non-monotonic trends in Appendix~\ref{sec:non_monotonic}.

\begin{table}[t]
\centering
\scalebox{0.8}{
\begin{tabular}{cccc|cc}
\toprule
\multirow{2}{*}{\textbf{Model}} & \multicolumn{3}{c|}{\textbf{Paradigm}} & \multicolumn{2}{c}{\textbf{Statistic}} \\
\cmidrule(lr){2-6}
 & \textbf{ZS} & \textbf{FS} & \textbf{CoT} & \textbf{Avg.} & \textbf{Var.} \\
\midrule
Claude-Sonnet-4-Thinking & 85.27 & 85.60 & 86.87 & 85.91 & 0.48 \\
DeepSeek-R1              & 91.23 & 89.33 & 88.43 & 89.67 & 1.36 \\
Doubao-1.5-Thinking-Pro  & 93.67 & 90.63 & 91.73 & 92.01 & 1.58 \\
\bottomrule
\end{tabular}
}
\caption{Capability under three prompting paradigms.}
\label{tab:prompting-paradigms-top-models}
\end{table}

\paragraph{Finding 2: Models demonstrate significant domain-specific performance variations, with specialized ``thinking'' abilities offering only marginal gains.}
As shown in Table~\ref{tab:overall-ranking}, all models excel in Management and Economics but consistently underperform in Literature, Medicine, and Military, revealing persistent domain-specific knowledge gaps. Dedicated ``thinking'' modes yield only modest gains (e.g., Claude-Sonnet-4-Thinking exceeds its base by $\sim$0.03 points), suggesting that domain coverage rather than reasoning mode remains the dominant performance driver.

\paragraph{Finding 3: In dynamic knowledge-intensive evaluations, prompting paradigms have minimal impact, whereas external augmentation may boost performance.}
As shown in Table~\ref{tab:prompting-paradigms-top-models}, performance varies within three points across Zero-Shot, Few-Shot, and Chain-of-Thought prompting for all models (full results in Appendix~\ref{sec:Leaderboard}). By contrast, enabling web search boosts GPT-4o by 1.23 points (82.51 to 83.74), with the largest gains in Medicine and Literature, suggesting that external knowledge access is more impactful than prompt engineering for knowledge-intensive tasks.

\paragraph{Finding 4: Systematic error analysis reveals that disciplinary knowledge gaps and comprehension failures are the primary limitations of current models.}
Disciplinary knowledge gaps (47.7\%) and misunderstanding errors (39.8\%) together account for nearly 90\% of all failures, indicating that knowledge coverage and contextual understanding are the primary bottlenecks. Figure~\ref{fig:Failure Cases} shows representative cases. The error taxonomy and coding methodology are detailed in Appendix~\ref{sec:error_categorization}.

\begin{figure*}[t]
  \centering
  \includegraphics[width=0.9\textwidth]{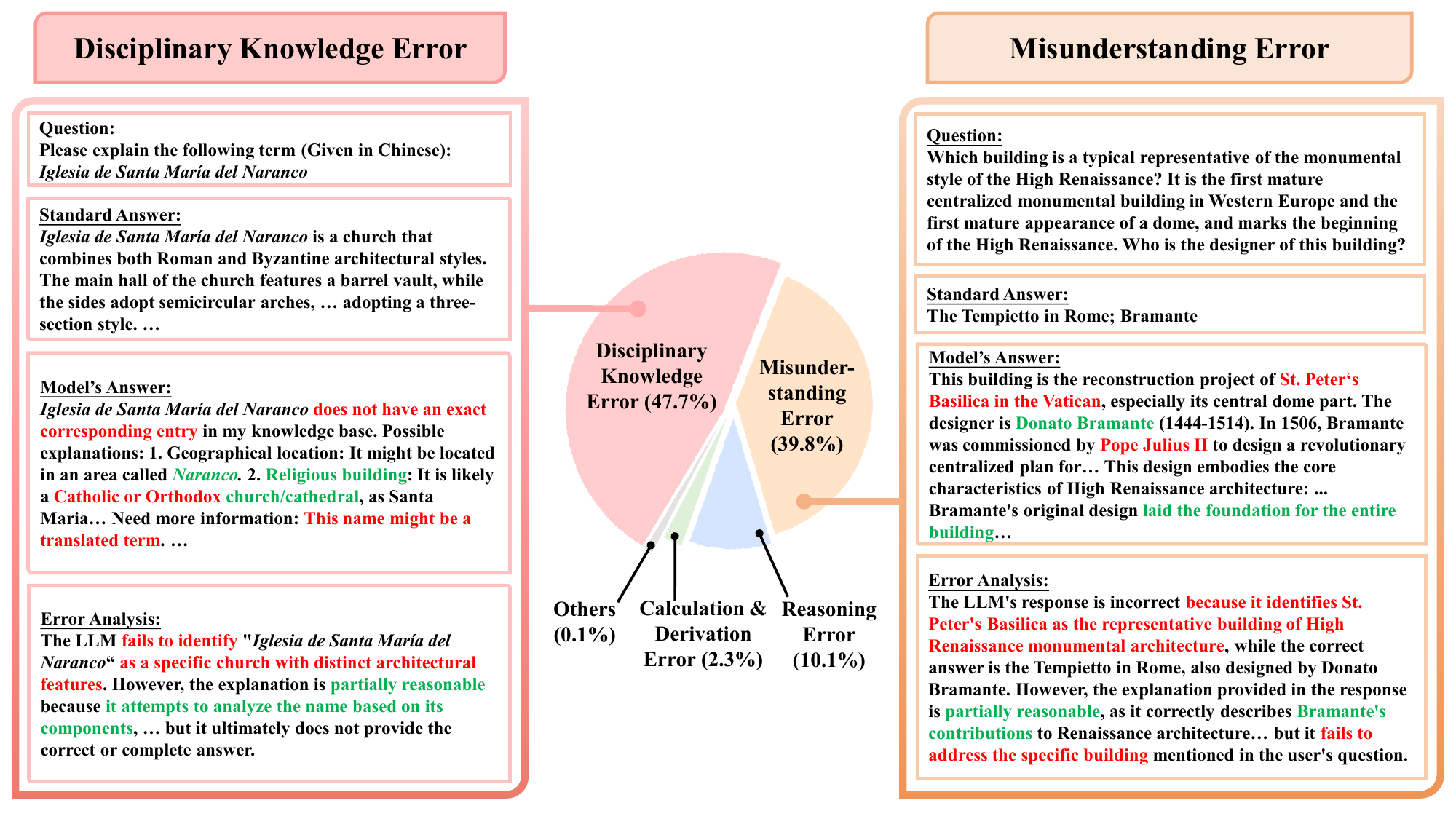}
  \caption{Distribution of model error causes and illustrative cases of the two most prevalent error types.}
  \label{fig:Failure Cases}
\end{figure*}

\subsection{Research Question II}

\paragraph{Finding 5: Dynamic rankings differ from static benchmarks, revealing contamination-driven distortions.}
To quantify the discrepancy between dynamic and static evaluation, we compare the model rankings from LLMEval-Fair with two representative static benchmarks, AGIEval \cite{AGIEval} and C-Eval \cite{c-eval}. As shown in Table~\ref{tab:static-benchmark-correlation}, the rank correlations are moderate ($\rho \approx 0.65\text{--}0.72$), indicating that static benchmark rankings do not consistently align with those produced by our dynamic evaluation.

Table~\ref{tab:rank-displacement} further reports per-model rank displacement ($\Delta\text{rank} = \text{Rank}_{\text{static}} - \text{Rank}_{\text{ours}}$). The two models most over-ranked by C-Eval (Claude-Sonnet-4, $\Delta=-2$; Doubao-1.5-Pro, $\Delta=-2$) are precisely those with the highest fill-in-the-blank completion rates (Table~\ref{tab:leakage_fill_compare}), linking rank distortions to contamination. Overall, 26.7\% of model pairs exhibit rank inversions against C-Eval versus only 13.3\% against AGIEval.
 
\begin{table}[t]

  \centering
  \footnotesize              
  \setlength{\tabcolsep}{4pt}  
  \begin{tabular}{lcc}
    \toprule
    Benchmark & Spearman $\rho$ & $p$-value \\
    \midrule
    AGIEval (EN) & 0.714 & 0.111 \\
    AGIEval (ZH) & 0.657 & 0.156 \\
    C-Eval       & 0.657 & 0.156 \\
    \bottomrule
  \end{tabular}
\caption{Rank correlation between LLMEval-Fair and static benchmarks.}
\label{tab:static-benchmark-correlation}
\end{table}

\begin{table}[t]
  \centering
  \footnotesize
  \setlength{\tabcolsep}{4pt}
  \begin{tabular}{l c cc cc}
    \toprule
    \multirow{2}{*}{\textbf{Model}} & \multirow{2}{*}{\textbf{Ours}} & \multicolumn{2}{c}{\textbf{C-Eval}} & \multicolumn{2}{c}{\textbf{AGIEval}} \\
    \cmidrule(lr){3-4} \cmidrule(lr){5-6}
     &  & Rank & $\Delta$ & Rank & $\Delta$ \\
    \midrule
    Gemini-2.5-Pro   & 1 & 2 & $+$1 & 1 & 0 \\
    DeepSeek-V3      & 2 & 4 & $+$2 & 3 & $+$1 \\
    Doubao-1.5-Pro   & 3 & 1 & $-$2 & 2 & $-$1 \\
    Qwen3-32B        & 4 & 5 & $+$1 & 5 & $+$1 \\
    Claude-Sonnet-4  & 5 & 3 & $-$2 & 4 & $-$1 \\
    GPT-4o           & 6 & 6 & \phantom{$+$}0 & 6 & \phantom{$+$}0 \\
    \bottomrule
  \end{tabular}
  \caption{Per-model rank displacement ($\Delta = \text{Rank}_\text{static} - \text{Rank}_\text{ours}$). Negative $\Delta$ indicates over-ranking by the static benchmark.}
  \label{tab:rank-displacement}
\end{table}

\paragraph{Finding 6: Static Benchmarks Suffer from Severe Data Contamination.}
We conduct fill-in-the-blank replay tests (1,000 questions, three attempts each) on AGIEval and C-Eval. As shown in Table~\ref{tab:leakage_fill_compare}, public benchmarks yield substantially higher completion counts than our private dataset, confirming that static benchmarks suffer from significant leakage while our question bank remains contamination-resistant.

\begin{table}[t]
  \centering
  \scalebox{0.8}{
    \begin{tabular}{lcccc}
      \toprule
      Model & AGI (EN) & AGI (ZH) & C-Eval & Ours \\
      \midrule
      DeepSeek-V3    & 97  & 153 & 136 & 80 \\
      ClaudeSonnet-4 & 179 & 248 & 224 & 179 \\
      Doubao-1.5     & 66  & 105 & 117 & 76 \\
      o3-mini        & 58  & 74  & 45  & 75 \\
      GPT-4o         & 54  & 86  & 62  & 48 \\
      Qwen3-32B      & 55  & 77  & 72  & 36 \\
      \bottomrule
    \end{tabular}
  }
  \caption{Comparison of successful fill-in completions for different models on static benchmarks and LLMEval-Fair.}
  \label{tab:leakage_fill_compare}
\end{table}

\subsection{Research Question III}
\label{Research Question III}
\paragraph{Finding 7: The relative ranking system demonstrates exceptional stability, with negligible variance across multi-round resampling and varying sample sizes.}
Multi-round resampling (n=1000, 2000, 4000) confirms that model ranking order remains identical across all runs, with relative scoring exhibiting negligible variance ($\sigma^2 \leq 1.68$). Full results are in Appendix~\ref{sec:detailsInExperiment}.

\begin{figure*}[t] 
  \centering
  
  \begin{minipage}[t]{0.48\textwidth}
    \centering
    \includegraphics[width=\linewidth]{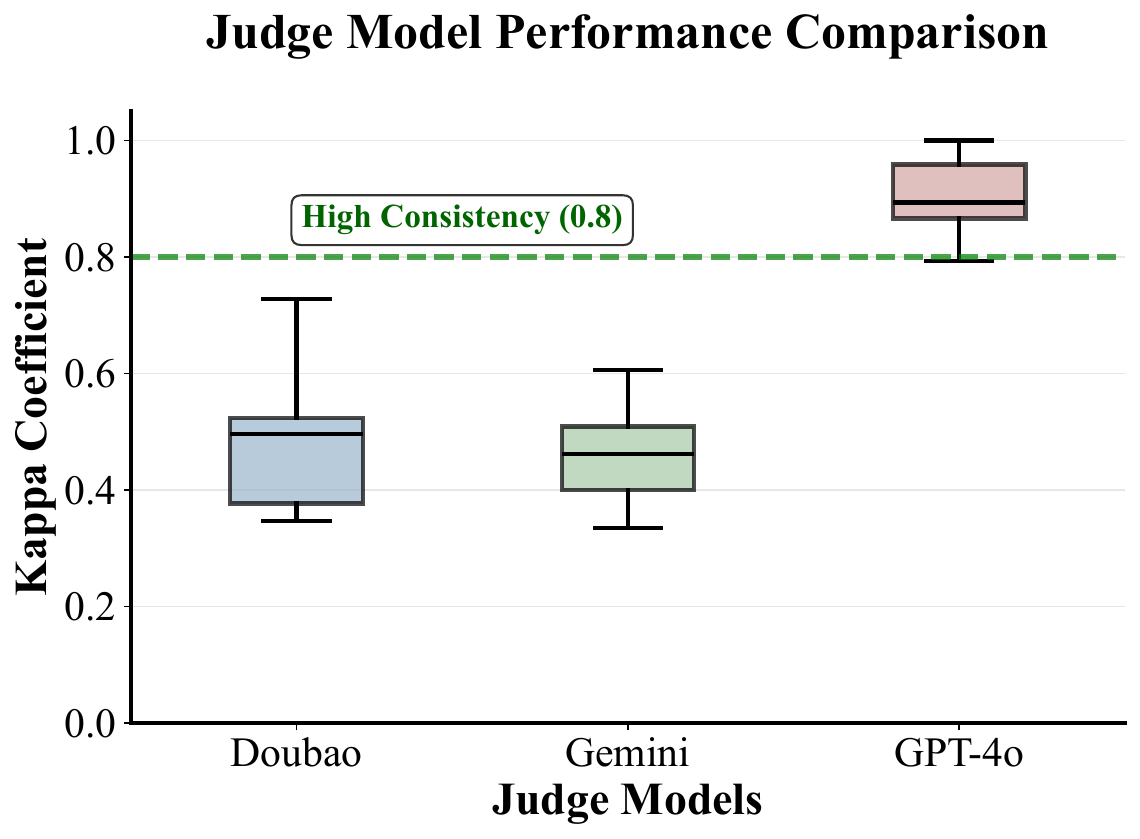}
    \caption{Cohen's $\kappa$ coefficients measuring agreement between human evaluators and three LLM judges across evaluations. GPT-4o achieves almost perfect agreement with human judgments ($\kappa = 0.907$).}
    \label{fig:kappa}
  \end{minipage}
  \hfill
  \begin{minipage}[t]{0.48\textwidth}
    \centering
    \includegraphics[width=\linewidth]{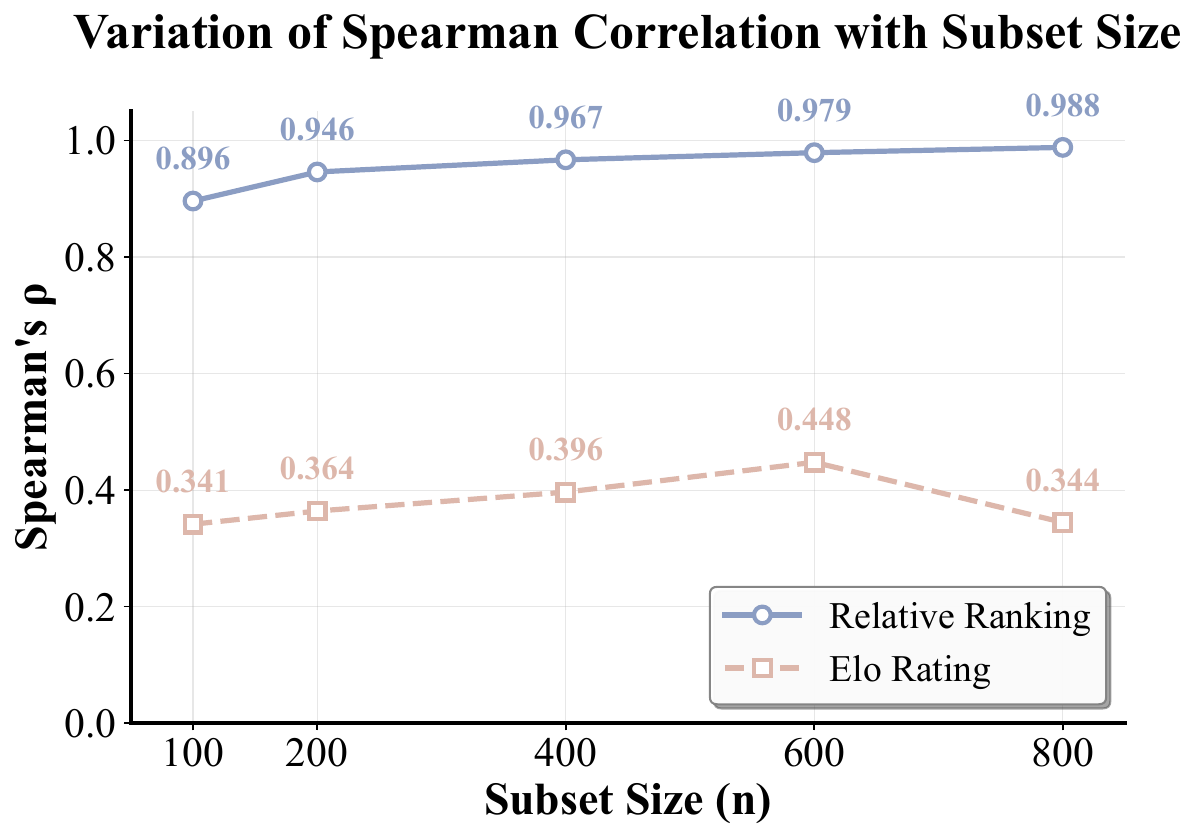}
    \caption{Relative ranking consistently outperforms Elo, reaching near-perfect correlation as the subset grows.}
    \label{fig:ablation-plot}
  \end{minipage}
  
\end{figure*}

\paragraph{Finding 8: The relative ranking system via LLM-as-Judge achieves high human-machine agreement, is robust to judge selection, and outperforms alternative ranking systems.}
We validate our ranking system on three dimensions. First, Cohen's $\kappa$ between LLM-as-a-Judge and human experts reaches 0.907 (Figure~\ref{fig:kappa}), indicating near-perfect agreement. Second, a cross-judge experiment with GPT-4o, Doubao-1.5-Thinking-Pro, and Gemini-2.5-Pro scoring the same 1,300 samples shows high ranking consistency (Spearman $\rho \geq 0.847$, all $p < 0.001$) and no family bias ($|\Delta| \leq 0.12$ on a 0--3 scale); full results are in Appendix~\ref{sec:cross_judge}. Third, our relative ranking consistently outperforms an Elo-style baseline in Spearman correlation and stability across sample sizes (Figure~\ref{fig:ablation-plot}).

\section{Related Work}

\textbf{LLM Benchmarks.}
Early benchmarks evaluate LLMs on fixed question sets spanning factual knowledge and reasoning \cite{mmlu,c-eval,AGIEval}, and the LLMEval series \cite{zhang2023llmeval,llmeval:zhang2024,Zhang2025LLMEvalMed} extends this paradigm to Chinese academic domains. To go beyond static tests, human-preference platforms leverage pairwise comparisons to rank models on conversational quality \cite{AlpacaEval,lmsys:zheng2024,chatbotarena:chiang2024,zhang2025two}, though they often lack depth in domain-specific reasoning and rely on non-expert annotators prone to stylistic bias \cite{humanprefer:raju2024}. More recently, a growing body of work probes specific capabilities, including self-knowledge \cite{yin-etal-2023-large}, hallucination detection \cite{sun-etal-2024-benchmarking}, learning efficiency \cite{Dou2025EvaLearn}, instruction following \cite{ye2025multi}, tool learning \cite{ye-etal-2024-toolsword,ye-etal-2024-rotbench,ye-etal-2025-toolhop,ye-etal-2025-tooleyes,ye2026cctu}, test-time scaling \cite{yin2025arise}, adaptive evaluation \cite{ding2025autojudger}, embodied interaction \cite{yang2025vehicleworld}, context learning \cite{Dou2026CLBench}, taxonomy-guided research \cite{Zhang2026TaxoBench}, agentic coding \cite{Ding2026OctoBench}, scientific workflows \cite{Shen2026SciAgentGym}, and long-horizon active interaction \cite{xu2026odysseyarena}. Despite this breadth, most benchmarks remain static and thus vulnerable to contamination.

\textbf{Benchmark Contamination.}
The static nature of fixed benchmarks exposes them to data leakage, inflating scores and misrepresenting true capabilities \cite{DBLP:journals/corr/abs-2412-03597,leakage:xu2024,DBLP:conf/naacl/Deng0TGC24}, and encourages overfitting where models memorize answers rather than generalize \cite{leakage:deng2024}. These reliability concerns have prompted growing interest in dynamic evaluation protocols that mitigate contamination through private question banks and resampling strategies, which is the central motivation of our work.

\textbf{LLM-as-a-Judge.}
Employing LLMs as scalable evaluators offers high throughput and strong human-preference alignment \cite{geval:liu2023,mtbench101}, yet systematic judge calibration and bias mitigation remain underexplored \cite{judging:zheng2023}. Our work addresses this gap through comprehensive cross-judge validation and human-machine agreement studies, demonstrating that a single well-calibrated judge can achieve near-perfect ranking consistency with human experts.

\section{Conclusion}
We introduced LLMEval-Fair, a dynamic, contamination-resistant evaluation framework built on a private 220k-question bank, a two-layer anti-cheating architecture, and an LLM-as-Judge relative ranking pipeline. A 30-month longitudinal study of nearly 60 open-source and proprietary models revealed a consistent performance ceiling near 90\%, systematic gaps in literature, medicine, and military knowledge, and widespread data leakage in static benchmarks. Our relative ranking method demonstrated negligible variance under multi-round resampling with varying sample sizes and achieved near-perfect agreement with human experts. We further confirmed that prompting format has minimal impact on performance in knowledge-intensive tasks, underscoring the superiority of dynamic, contamination-resistant evaluation over static benchmarks and the need for more trustworthy benchmarking practices.

\section*{Limitations}
The current benchmark is limited to Chinese-language academic questions, although the evaluation framework itself (dynamic sampling, anti-cheating architecture, relative ranking) is language-agnostic and applicable to any language given an appropriate question bank. Additionally, the sheer size of the question bank (over 220,000 items) makes comprehensive evaluation resource-intensive, requiring substantial compute, time, and human effort for large-scale inference, result validation, and ongoing dataset maintenance.

To facilitate reproducibility and community adoption, we have open-sourced the full 220k question bank, all evaluation results, a detailed data schema, recommended sampling strategies, and ethical usage guidelines at \url{https://github.com/llmeval/LLMEval-Fair}.

\bibliography{main.bib}

\clearpage
\appendix
\section*{\centering \LARGE{Appendix}}
\section{Dataset}
\label{sec:Dataset}

This section provides supplementary information on our LLMEval-Fair dataset employed in the study, to clarify the academic disciplines and question types covered, and to elaborate methodologies for scaling the questions.

\subsection{Categories of Academic Displines}

\subsubsection{Categories of Academic Disciplines}
\begin{figure*}[hbtp]
\centering
\includegraphics[width=0.8\linewidth]{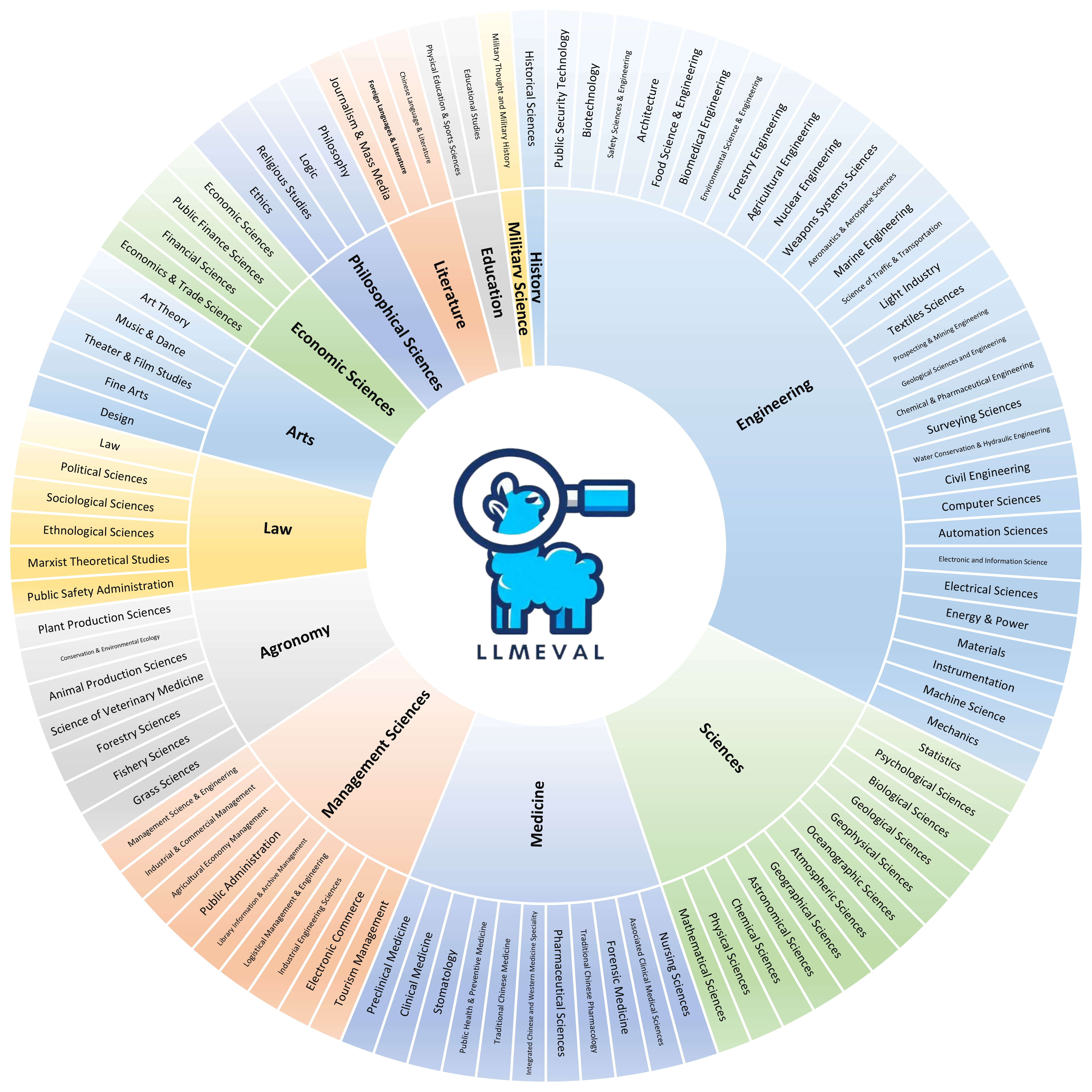}
\caption{Categories of Primary and Secondary Academic Disciplines.}
\label{fig:Categories of academic Disciplines}
\end{figure*}

As illustrated in Figure~\ref{fig:Categories of academic Disciplines}, we collected graduate-level examination questions spanning 13 primary (Philosophical Sciences, Economic Sciences, Law, Education, Literature, History, Engineering, Agronomy, Medicine, Military Science, Management Sciences, Arts, and Sciences) and more than 50 secondary academic disciplines recognized by China's Ministry of Education. Two-thirds of the questions are derived from Chinese universities' Postgraduate Entrance Exams, and one-third are from Undergraduate Final Exams of comparable difficulty. Detailed distribution of question sources is listed in Table~\ref{tab:question_type_distribution}.

\subsubsection{Categories of Question Types}

The raw dataset encompasses a diverse range of original question formats, including multiple-choices, fill-in-the-blank, true-or-false, short-answer, term explanation, and material analysis questions. Following question expansion and formatting, we have unified all questions and their answers into a fill-in-the-blank-like question-answer format, abandoning the complex answer structures of various question types, such as options A to E in multiple-choice questions, and ``true" or ``false" in true-or-false questions. This natural question-answer format enables the dataset to more thoroughly showcase the model's capabilities.

\begin{table*}[htbp]
    \centering
    \begin{tabular}{lcc}
        \hline
        \textbf{Type} & \textbf{Number of Topics} & \textbf{Proportion (\%)} \\
        \hline
        Undergraduate Final Exams & 26633 & 34.1 \\
        Postgraduate Entrance Exams & 51376 & 65.9 \\
        \textbf{Total} & \textbf{78009} & 100.0 \\
        \hline
        \hline
        Undergraduate Final Exams & 71038 & 31.1 \\
        Postgraduate Entrance Exams & 157566 & 68.9 \\
        \textbf{Total} & \textbf{228604} & 100.0 \\
        \hline
        \hline
    \end{tabular}
    \caption{Distribution of question number and proportions for Undergraduate Final Exams and Postgraduate Entrance Exams.}
    \label{tab:question_type_distribution}
\end{table*}

\subsection{Details of Expanding the Dataset}

As shown in Table~\ref{tab:discipline_counts}, we have amassed a substantially large quantity of original questions, with a marked surge in numbers following the expansion of our dataset.

\begin{table*}[htbp]
    \centering
    \scalebox{0.9}{ 
    \begin{tabular}{lcccc}
        \toprule
        \textbf{Subject} & \textbf{Original Count} & \textbf{Rewrited Count} & \textbf{Increase Count} & \textbf{Increase Percentage (\%)} \\
        \midrule
        Philosophical Sciences & 2194 & 10969 & 8775 & 399.95 \\
        Medicine & 30772 & 109974 & 79202 & 257.38 \\
        Law & 9262 & 30116 & 20854 & 225.16 \\
        Management Sciences & 2448 & 7945 & 5497 & 224.55 \\
        Engineering & 4926 & 13263 & 8337 & 169.24 \\
        Sciences & 6182 & 15669 & 9487 & 153.46 \\
        Economic Sciences & 9245 & 18124 & 8879 & 96.04 \\
        Military Science & 611 & 1187 & 576 & 94.27 \\
        Education & 2781 & 5094 & 2313 & 83.17 \\
        History & 1749 & 3178 & 1429 & 81.70 \\
        Literature & 7839 & 13085 & 5246 & 66.92 \\
        \midrule
        \textbf{Total} & \textbf{78009} & \textbf{228604} & \textbf{150595} & \textbf{193.05} \\
        \bottomrule
    \end{tabular}
    }
    \caption{Distribution of Original and Rewritten Counts Across Disciplines.}
    \label{tab:discipline_counts}
\end{table*}

\subsubsection{Original Data Construction Pipeline}

Original questions, structured simply and organized by subject for initial collation, follow this construction pipeline: first, converting Excel, Word, and PDF test papers to TXT; then batch splitting into JSON-formatted questions via scripts; and finally conducting data screening.

The latter involves three steps: (1) Expert review removes factually erroneous or irrelevant questions. (2) Batch splitting classifies and isolates questions, addressing errors such as content overlap, missing questions, or misclassifications. (3) Format cleaning resolves encoding conflicts, special characters, symbol consistency, redundancy, and typos.

\subsubsection{Data Expanding Pipeline}

To accommodate diverse application scenarios, this study proposes an augmented data format that complements the original question structure, The augmented dataset incorporates comprehensive metadata, including primary disciplinary categories, secondary disciplinary categories, question descriptions, answer content, and unique identifiers (UUIDs), which facilitates categorized data management. An example of expanding the Multiple-Choices question is shown in Figure~\ref{fig:expanding_multiple_choice}.

Two additional key verification procedures are implemented upon the core augmentation strategy: (1) Format validation, which entails checking the consistency of option counts for multiple-choice questions and the alignment of answer spaces for Fill-in-the-Blank questions; and (2) Redundancy checks, which involve detecting duplicates among split questions and ensuring the uniqueness of question UUIDs. The data formats before and after expanding is illustrated in Figure~\ref{fig:expanded_format}.

\begin{figure*}[ht]
    \centering
\begin{tcolorbox}[colback=gray!5!white,colframe=gray!75!black]

\texttt{Original Multiple-Choices questions}:\\

\textbf{Title}: It is known that the sixth level of a complete binary tree (let the root be the first level) has eight leaves, then the number of nodes of the complete binary tree is at most (    ).\\
A. 39\\
B. 52\\
C. 111\\ 
D. 119\\

\textbf{Answer:} C\\

\texttt{Expanded questions}:\\

\textbf{Question 1:} It is known that the sixth level of a complete binary tree (let the root be the first level) has eight leaves, then the number of nodes of the complete binary tree is at most: (    )\\
Is it correct to place the answer ``39" in the provided space?\\
\textbf{Answer:} False\\

\textbf{Question 2:} It is known that the sixth level of a complete binary tree (let the root be the first level) has eight leaves, then the number of nodes of the complete binary tree is at most: (    )\\
Is it correct to place the answer ``52" in the provided space?\\
\textbf{Answer:} False\\

\textbf{Question 3:} It is known that the sixth level of a complete binary tree (let the root be the first level) has eight leaves, then the number of nodes of the complete binary tree is at most: (    )\\
Is it correct to place the answer ``111" in the provided space?\\
\textbf{Answer:} True\\

\textbf{Question 4:} It is known that the sixth level of a complete binary tree (let the root be the first level) has eight leaves, then the number of nodes of the complete binary tree is at most: (    )\\
Is it correct to place the answer ``119" in the provided space?\\
\textbf{Answer:} False\\
  
\end{tcolorbox}
    \caption{
    Example of how to expand a Multiple Choice question.
    }
\label{fig:expanding_multiple_choice}
\end{figure*}

\begin{figure*}[ht]
    \centering
\begin{tcolorbox}[colback=gray!5!white,colframe=gray!75!black]

  \texttt{category}: Engineering,\\
  \texttt{sub\_category}: Computer Science,\\
  \texttt{question\_uuid}: a4b92619-f0ac-xxxx-xxxx-9f11ce3b4704,\\\
  \texttt{dialog}: \\
    \texttt{role}: user,\\
    \texttt{content}: Judge the correctness of the following statements, answer true/false, and give your reasons. The continue statement in a loop breaks out of the current loop.\\
  \texttt{hint}: false,\\
  
  \texttt{gpt4res}: Answer: false. The reason is that the continue statement does not break out of the entire loop.It skips the rest of the current loop and immediately moves on to the next iteration of the loop instead of ending the current loop completely.\\
  \texttt{gpt4judge}: \\
    \texttt{judge}: Overall Rating: 3  The reason why you gave this rating: The LLM provided the correct answer (false) and gave a reasonable explanation. It accurately explained that the `continue` statement does not terminate the entire loop but instead skips the remaining part of the current iteration and proceeds to the next iteration of the loop. This explanation aligns with the correct behavior of the `continue` statement in programming.,\\
   \texttt{stars}: 3,\\
   \texttt{reason}: The LLM provided the correct answer (false) and gave a reasonable explanation. It accurately explained that the `continue` statement does not terminate the entire loop but instead skips the remaining part of the current iteration and proceeds to the next iteration of the loop. This explanation aligns with the correct behavior of the `continue` statement in programming.\\
  
\end{tcolorbox}
    \caption{
    Detailed entries of a single question after expanding.
    }
\label{fig:expanded_format}
\end{figure*}

\section{Prompts}
\label{sec:prompts}

This section presents the complete set of prompts used in LLMEval-Fair for different evaluation paradigms. We provide the specific prompt templates for few-shot learning, chain-of-thought reasoning, and LLM-based automated evaluation to ensure reproducibility of our experimental results.

The prompts used for testing few-shot and chain-of-thought methods are shown in Figures \ref{fig: prompt for few-shot} and \ref{fig: prompt for cot}, respectively. The prompt used for LLM-based evaluation is shown in Figure \ref{fig: prompt for judge}.

\begin{figure*}[ht]
    \small
    \centering
\begin{tcolorbox}[colback=gray!5!white,colframe=gray!75!black,fontupper=\footnotesize,]
\textbf{Input:}
\\~\\Here are several examples: \\
 \\
Question: \\
Please determine the correctness of the following statement. Answer with true/false and provide a reason. If all elements below the diagonal in the adjacency matrix of a directed graph are zero, then the graph must have a topological ordering. \\
Answer: \\
true. Because if all elements below the diagonal in the adjacency matrix of a directed graph are zero, the graph is a Directed Acyclic Graph (DAG), so it must have a topological ordering. \\
 \\
Question: \\
Please explain the following term: ``Double Hundred Policy". \\
Answer: \\
It refers to ``let a hundred flowers bloom, let a hundred schools of thought contend." This was a policy on science and culture officially proposed by Mao Zedong in 1956 and confirmed by the Central Committee of the Communist Party of China. The policy was severely undermined after 1957, but was reestablished and implemented following the Third Plenary Session of the 11th Central Committee. \\
 \\
Question: \\
Please determine the correctness of the following statement. Answer with true/false and provide a reason. According to the convertibility theory, the scope of commercial banks’ assets expanded from short-term turnover loans to consumer loans. \\
Answer: \\
false. Convertibility Theory: also known as the asset conversion theory. This theory suggests that to maintain liquidity for withdrawals, commercial banks can invest part of their funds in transferable securities. Since these profitable assets can be sold at any time and converted into cash, loans are not necessarily limited to short-term and self-liquidating types. Clearly, this theory emerged in the context of developing financial instruments and markets. Significance: it expanded the scope of bank asset operations. Drawbacks: it does not guarantee that assets can be liquidated without capital loss (which requires high asset quality and stable market conditions); it is also constrained by central bank monetary policy (e.g., the risk of a rise in discount rates). This theory provides a theoretical basis for Chinese commercial banks to engage in securities business (investment operations). However, in China, commercial banks' investment activities are restricted due to: 1. limited investment instruments and underdeveloped credit mechanisms; 2. management systems narrowing investment scopes, and separation of operations preventing commercial banks from investing in stocks; 3. the nature of state-owned commercial banks limits their willingness for autonomous investment. \\
 \\
Question: \\
Why can the results of animal experiments not be fully applied to clinical practice? \\
Answer: \\
Because there are differences between humans and animals not only in cellular morphology and metabolism, but also due to the highly developed human nervous system, which is associated with language and thought (the second signaling system). Although there are similarities, the essential differences mean that human diseases cannot all be replicated in animals. Even if they can be replicated, animal responses are simpler than human responses. Therefore, results from animal experiments cannot be mechanically and fully applied to clinical practice without analysis. Only by comparing, analyzing, and synthesizing animal experiment results with clinical data can they be used as references in clinical medicine and provide a basis for studying the causes, mechanisms, prevention, and treatment of clinical diseases. \\
 \\
The following is the question to be answered: \\
Question:\texttt{\{question\}} \\
Answer: \\

\end{tcolorbox}
    \caption{
    The Few-shot Prompt Template.
    }
\label{fig: prompt for few-shot}
\end{figure*}

\begin{figure*}[ht]
    \centering
\begin{tcolorbox}[colback=gray!5!white,colframe=gray!75!black]
\textbf{Input:}
\\~\\
\texttt{\{question\}} \\
Please think step by step and provide the final answer.\\

\end{tcolorbox}
    \caption{
    The Chain-of-Thought Prompt Template.
    }
\label{fig: prompt for cot}
\end{figure*}

\begin{figure*}[ht]
    \centering
\begin{tcolorbox}[colback=gray!5!white,colframe=gray!75!black]
\textbf{Input:}
\\~\\
Please evaluate the following response from the \\
LLM regarding a discipline-specific question based  \\
on the following criteria. You must score it on a scale of 0, 1, 2 or 3 \\
stars: \\
 \\
Overall Rating: \\
0 star indicates wrong answer with a wrong explanation \\
1 stars indicate wrong answer but a partially reasonable explanation \\
2 stars indicate a correct answer with a partially reasonable explanation \\
3 stars indicate an correct answer with a reasonable explanation \\
 \\
User: \texttt{\{question\}} \\
 \\
LLM: \texttt{\{LLM response\}} \\
 \\
The correct answer to user’s question is: \texttt{\{correct answer\}} \\
 \\
You must provide your feedback in the following format: \\
``Overall Rating":numbers of its stars(int) \\
The reason why you gave this rating: <Your Reason>(str)' \\

\end{tcolorbox}
    \caption{
    The Prompt Template for LLM Judgement.
    }
\label{fig: prompt for judge}
\end{figure*}

\section{Evaluation Model Selection}
\label{app:model_selection}

To ensure a comprehensive and rigorous assessment, we conducted a preliminary evaluation on a broad set of 59 large language models. From this extensive pool, we selected a representative subset of 17 models for the detailed analysis presented in the main text. The detailed release dates for these selected models are provided in Table~\ref{model_dates}.

The proprietary frontier models include the \textbf{OpenAI} lineup, featuring the widely deployed \textbf{GPT-4o} and \textbf{GPT-4o-search} \cite{OpenAI2023GPT4}, alongside the next-generation flagship \textbf{GPT-5} \cite{openai2025gpt5}. Furthermore, we assess the \textbf{o1} series \cite{openai2024o1} and the efficient \textbf{o3-mini} \cite{openai2025o3mini}, which represent specialized reasoning models trained with large-scale reinforcement learning to solve complex scientific and mathematical problems.

We include models capable of extended reasoning from other major providers. \textbf{Anthropic} is represented by the standard \textbf{Claude-Sonnet-4} \cite{anthropic2024claude4}, as well as the \textbf{Claude-Sonnet-4-Thinking} and the advanced \textbf{4.5-Thinking} variants \cite{anthropic2025claude45}, which operate in a specialized mode to perform self-reflection before generating responses. Similarly, \textbf{Google}'s contribution consists of \textbf{Gemini-2.5-Pro} \cite{google2025gemini25} and its reasoning-enhanced counterpart, \textbf{Gemini-2.5-Pro-Thinking} \cite{google2025gemini25}, which incorporates internal chain-of-thought processes.

Regarding open-weights architectures and diverse scaling strategies, the study includes the \textbf{DeepSeek} family: \textbf{DeepSeek-V3} \cite{deepseek2025v3}, a strong Mixture-of-Experts (MoE) model, and \textbf{DeepSeek-R1} \cite{deepseek2025r1}, a model specifically optimized for reasoning tasks through post-training. The \textbf{Qwen-3} series is also evaluated to represent distinct points on the parameter spectrum, featuring both the massive \textbf{235B} model and the compact \textbf{32B} variant \cite{Qwen3}.

Finally, we evaluate other high-performing systems to ensure a broad representation of the current landscape. This includes \textbf{Moonshot}'s \textbf{Kimi-K2} \cite{moonshot2025kimik2} and the \textbf{Doubao-1.5} series. Notably, \textbf{Doubao-1.5-Thinking-Pro} \cite{bytedance2025doubao} serves as a key reference in our study, having demonstrated state-of-the-art capabilities in preliminary screenings.

\begin{table*}[t]
\centering

\footnotesize 
\setlength{\tabcolsep}{4pt} 
\renewcommand{\arraystretch}{1.2} 

\begin{tabular}{ll @{\hskip 0.5cm} ll} 
    \toprule
    \textbf{Model} & \textbf{Released Date} & \textbf{Model} & \textbf{Released Date} \\
    \midrule
    
    Doubao-1.5-Thinking-Pro \citeyearpar{bytedance2025doubao} & April 15, 2025 & 
    Claude-Sonnet-4.5-Thinking \citeyearpar{anthropic2025claude45} & September 29, 2025 \\
    
    Doubao-1.5-Pro \citeyearpar{bytedance2025doubao} & January 15, 2025 & 
    Claude-Sonnet-4-Thinking \citeyearpar{anthropic2024claude4} & May 14, 2025 \\
    
    Gemini-2.5-Pro-Thinking \citeyearpar{google2025gemini25} & June 5, 2025 & 
    Claude-Sonnet-4 \citeyearpar{anthropic2024claude4} & May 14, 2025 \\
    
    Gemini-2.5-Pro \citeyearpar{google2025gemini25} & June 5, 2025 & 
    o1 \citeyearpar{openai2024o1} & December 17, 2024 \\
    
    DeepSeek-R1 \citeyearpar{deepseek2025r1} & May 28, 2025 & 
    o3-mini \citeyearpar{openai2025o3mini} & January 29, 2025 \\
    
    DeepSeek-V3 \citeyearpar{deepseek2025v3} & March 24, 2024 & 
    GPT-4o-search \citeyearpar{OpenAI2023GPT4} & November 20, 2024 \\
    
    Qwen-3-235B \citeyearpar{Qwen3} & April 29, 2025 & 
    GPT-4o \citeyearpar{OpenAI2023GPT4} & November 20, 2024 \\
    
    Qwen-3-32B \citeyearpar{Qwen3} & April 29, 2025 & 
    Kimi-K2 \citeyearpar{moonshot2025kimik2} & September 5, 2025 \\
    
    GPT-5 \citeyearpar{openai2025gpt5} & June 7, 2025 & 
    -- & -- \\ 
    
    \bottomrule
\end{tabular}
\caption{The representative list of models selected from a total of 59 evaluated LLMs. Dates for unreleased models are estimated based on technical previews.}
\label{model_dates}

\end{table*}

\section{The Elo Rating System}
\label{sec:elo}
The Elo rating system is a widely recognized method for calculating relative skill levels in zero-sum games, originally developed for chess and recently popularized for evaluating Large Language Models (e.g., Chatbot Arena). This system derives ratings from the outcomes of pairwise comparisons, providing a probabilistic framework to predict the likelihood of one entity outperforming another. Given two entities $A$ and $B$ with current ratings $R_A$ and $R_B$, the expected score $E_A$ (representing the probability of $A$ winning) is calculated using a logistic curve:

\begin{equation}
    E_A = \frac{1}{1 + 10^{(R_B - R_A) / 400}}
\end{equation}

Following a match, the ratings are updated based on the discrepancy between the actual outcome $S_A$ (where 1 represents a win, 0 a loss, and 0.5 a tie) and the expected probability. The update rule is given by $R_A' = R_A + K(S_A - E_A)$, where $K$ is a constant factor that determines the sensitivity of the rating adjustment.

\section{LLMEval-Fair Leaderboard}
\label{sec:Leaderboard}
This section presents comprehensive evaluation results from our longitudinal study tracking nearly 60 LLMs from late 2023 to mid-2025. We provide complete performance rankings and analyze the consistency of model capabilities across different prompting paradigms.

We tracked nearly 60 LLMs from late 2023 to mid-2025. Here, we present the complete evaluation results of our model assessments.
The comprehensive results, including scores for all models across 10 academic disciplines, are presented in Table~\ref{tab:overall-leaderboard}.

The models we selected in main paper was evaluated across three prompting paradigms: Zero-Shot (ZS), Few-Shot (FS), and Chain-of-Thought (CoT). As shown in Table~\ref{tab:prompting-paradigms-top-models}, the performance variance across these paradigms remains below 1.6 points for all evaluated models, indicating that core capabilities are not significantly influenced by the prompting format.

\begin{table*}[t]
  \centering  
  \footnotesize
  \scalebox{0.85}{%
  \begin{tabular}{lcccccccccccc}
    \hline
    Model & $R_{\text{SOTA}}^{\text{model}}$ & $S_{\text{model}}$ & Eng. & Econ. & Edu. & Law & Lit. & Mgmt. & Sci. & Hist. & Med. & Mil. \\
    \hline
    \multicolumn{13}{l}{\textit{Open-source LLMs}} \\
    DeepSeek-R1 \citeyearpar{deepseek2025r1}                   & \textbf{97.40} & \textbf{91.23} & \textbf{9.47} & 9.43 & \textbf{9.27} & 9.37 & \textbf{8.83} & 9.37 & \textbf{9.03} & \textbf{9.53} & 8.50 & 8.43 \\
    DeepSeek-V3 \citeyearpar{deepseek2025v3}                   & 96.47 & 90.36 & 9.30 & \textbf{9.57} & 8.93 & 9.23 & 8.60 & 9.13 & 8.97 & 9.47 & \textbf{8.83} & 8.33 \\
    Qwen3-235B \citeyearpar{Qwen3}                               & 96.42 & 90.32 & 9.23 & 9.43 & 9.03 & \textbf{9.50} & 8.23 & 9.43 & 8.97 & 9.17 & 8.73 & \textbf{8.60} \\
    QwQ-32B \citeyearpar{qwen_qwq_32b_2024}                       & 94.51 & 88.53 & 8.30 & 9.46 & 9.23 & 9.33 & 7.83 & 9.46 & 8.65 & 9.27 & 8.57 & 8.43 \\
    DeepSeek-V3.2 \citeyearpar{DeepSeek-V3.2}                 & 92.27 & 86.43 & 8.73 & 9.13 & 8.53 & 8.70 & 7.40 & 9.33 & 8.87 & 9.37 & 8.53 & 7.83 \\
    Qwen3-32B \citeyearpar{Qwen3}                                & 92.22 & 86.38 & 8.43 & 9.10 & 8.57 & 9.10 & 7.77 & \textbf{9.47} & 8.67 & 9.30 & 7.70 & 8.27 \\
    GLM-4-32B \citeyearpar{zai_glm_4_32b_0414_2026}                     & 88.43 & 82.83 & 7.77 & 8.97 & 8.33 & 8.33 & 7.03 & 9.13 & 8.27 & 8.77 & 8.23 & 8.00 \\
    Qwen2.5-32B-Instruct \citeyearpar{Qwen2.5}          & 85.06 & 79.68 & 7.70 & 8.57 & 8.33 & 8.33 & 6.70 & 8.50 & 8.17 & 7.70 & 7.60 & 8.08 \\
    Qwen-Turbo-1101 \citeyearpar{alibaba_qwen_turbo_1101_2024}               & 83.72 & 78.42 & 7.97 & 8.37 & 8.03 & 8.23 & 6.40 & 8.50 & 8.10 & 7.50 & 7.27 & 8.05 \\
    Yi-34B-Chat \citeyearpar{Yi}                   & 70.15 & 65.71 & 5.77 & 6.63 & 7.37 & 7.53 & 5.47 & 5.77 & 5.47 & 7.47 & 6.30 & 7.93 \\
    Megrez-3B-Instruct \citeyearpar{Megrez-Omni}            & 67.01 & 62.77 & 5.80 & 6.77 & 6.80 & 7.13 & 5.40 & 6.87 & 5.70 & 6.53 & 5.70 & 6.07 \\
    Qwen2-7B-Instruct \citeyearpar{Qwen2}             & 65.15 & 61.03 & 5.47 & 6.73 & 6.33 & 7.60 & 5.13 & 6.17 & 6.17 & 5.73 & 5.33 & 6.37 \\
    Nanbeige-Plus \citeyearpar{nanbeige_plus_chat_v01_2024}                 & 65.10 & 60.98 & 5.78 & 5.57 & 6.77 & 7.37 & 5.37 & 5.93 & 5.45 & 6.30 & 5.67 & 6.77 \\
    Phi-4-Final \citeyearpar{Phi-4}                   & 63.98 & 59.93 & 5.80 & 6.47 & 6.23 & 6.53 & 5.53 & 6.30 & 6.27 & 5.50 & 5.43 & 5.87 \\
    Llama-3.2-90B-Vision \citeyearpar{meta_llama_32_90b_vision_instruct_2024}          & 61.74 & 57.83 & 5.63 & 6.33 & 6.20 & 5.80 & 4.73 & 6.10 & 6.57 & 5.03 & 5.27 & 6.17 \\
    Llama-3.3-70B \citeyearpar{meta_llama_33_70b_instruct_2024}                 & 60.85 & 57.00 & 5.80 & 6.90 & 5.63 & 5.70 & 5.47 & 5.70 & 6.30 & 4.70 & 4.87 & 5.93 \\
    Baichuan2-13B-Chat \citeyearpar{Baichuan2}            & 58.28 & 54.59 & 4.47 & 5.53 & 7.40 & 6.90 & 4.63 & 4.80 & 4.33 & 6.23 & 4.60 & 5.70 \\
    Qwen-plus \citeyearpar{alibaba_qwen_plus_2026}                     & 56.58 & 53.00 & 4.40 & 5.10 & 6.53 & 6.53 & 5.00 & 4.77 & 4.87 & 5.17 & 5.13 & 5.50 \\
    Qwen-turbo \citeyearpar{alibaba_qwen_turbo_2026}                    & 55.76 & 52.23 & 4.10 & 6.07 & 6.63 & 6.43 & 4.43 & 4.53 & 4.97 & 5.27 & 4.37 & 5.43 \\
    Nanbeige-16B \citeyearpar{nanbeige_16b_2023}                  & 55.45 & 51.94 & 4.37 & 5.30 & 6.50 & 6.30 & 3.97 & 4.70 & 4.07 & 5.90 & 4.73 & 6.10 \\
    Mixtral-8x7B-Instruct \citeyearpar{Mixtral}         & 51.69 & 48.42 & 4.27 & 5.47 & 6.47 & 6.40 & 3.13 & 4.50 & 5.07 & 3.57 & 4.37 & 5.17 \\
    ChatGLM-2-6B \citeyearpar{ChatGLM}                  & 42.31 & 39.63 & 2.33 & 3.77 & 5.97 & 6.13 & 2.83 & 3.83 & 2.60 & 3.80 & 4.00 & 4.37 \\
    Llama-3.1-8B \citeyearpar{meta_llama_31_8b_2024}                  & 41.25 & 38.64 & 3.87 & 4.20 & 4.27 & 4.17 & 3.50 & 3.83 & 4.30 & 3.17 & 3.20 & 4.13 \\
    Ziya-13B-v1.1 \citeyearpar{Ziya}                 & 40.18 & 37.64 & 2.77 & 3.97 & 5.17 & 5.33 & 2.80 & 3.77 & 2.53 & 3.70 & 3.03 & 4.57 \\
    InternLM-7B-Chat \citeyearpar{internlm_chat_7b_2023}              & 38.71 & 36.26 & 2.63 & 3.67 & 4.87 & 5.57 & 3.17 & 3.33 & 2.33 & 4.03 & 3.13 & 3.53 \\
    Linly-LLaMA2-13B \citeyearpar{linly_chinese_llama2_13b_2023}              & 37.03 & 34.69 & 2.20 & 3.77 & 4.50 & 5.00 & 2.43 & 3.33 & 2.53 & 3.90 & 2.50 & 4.53 \\
    Phi-3-Medium-128K \citeyearpar{Phi-4}             & 36.95 & 34.61 & 2.27 & 4.17 & 3.70 & 4.23 & 2.87 & 4.50 & 3.57 & 3.20 & 2.27 & 3.83 \\
    BELLE-Llama2-13B-Chat \citeyearpar{belle_llama2_13b_chat_04m_2023}         & 36.25 & 33.96 & 2.57 & 3.07 & 4.93 & 4.73 & 2.83 & 3.80 & 2.43 & 3.33 & 2.40 & 3.87 \\
    Llama-2-7B-Chat \citeyearpar{Llama2}               & 25.22 & 23.62 & 1.53 & 3.43 & 3.00 & 3.73 & 1.73 & 2.43 & 1.97 & 2.17 & 0.80 & 2.83 \\
    \hline
    \multicolumn{13}{l}{\textit{Closed-source LLMs}} \\
    Doubao-1.5-Thinking-Pro \citeyearpar{bytedance2025doubao}   & \textbf{100.00} & \textbf{93.67} & \textbf{9.47} & \textbf{9.67} & \textbf{9.43} & \textbf{9.77} & \textbf{8.93} & 9.53 & \textbf{9.23} & \textbf{9.70} & \textbf{8.97} & \textbf{8.97} \\
    Gemini-2.5-Pro \citeyearpar{google2025gemini25}             & 97.22 & 91.07 & 9.20 & 9.47 & 9.20 & 9.30 & 8.43 & 9.63 & 9.07 & 9.40 & 8.50 & 8.87 \\
    Gemini-2.5-Pro-Thinking \citeyearpar{google2025gemini25}    & 97.15 & 91.00 & 9.13 & 9.50 & 9.37 & 9.47 & 8.40 & 9.63 & 9.20 & 9.27 & 8.30 & 8.73 \\
    Doubao-1.5-Pro \citeyearpar{bytedance2025doubao}            & 95.68 & 89.62 & 8.83 & 9.03 & 9.13 & 9.43 & 8.57 & 9.27 & 8.83 & 9.10 & 8.60 & 8.83 \\
    GLM-4.6 \citeyearpar{zhipu2025glm46}                         & 95.26 & 89.23 & 8.80 & 9.27 & 8.70 & 9.23 & 8.40 & \textbf{9.63} & 8.90 & 9.30 & 8.43 & 8.57 \\
    Kimi-K2 \citeyearpar{moonshot2025kimik2}                    & 94.27 & 88.30 & 9.23 & 9.17 & 8.80 & 9.00 & 8.40 & 9.17 & 8.77 & 9.13 & 8.53 & 8.10 \\
    GPT-5 \citeyearpar{openai2025gpt5}                          & 93.84 & 87.90 & 8.83 & 9.37 & 8.90 & 8.87 & 8.10 & 9.10 & 8.90 & 9.03 & 8.50 & 8.30 \\
    Claude-Sonnet-4.5-Thinking \citeyearpar{anthropic2025claude45} & 93.48 & 87.57 & 8.90 & 9.17 & 8.80 & 8.97 & 8.00 & 9.23 & 8.90 & 9.00 & 8.27 & 8.33 \\
    o1 \citeyearpar{openai2024o1}                               & 93.36 & 87.45 & 8.90 & 9.30 & 8.67 & 8.77 & 7.73 & 9.27 & 8.90 & 8.97 & 8.17 & 8.77 \\
    Claude-Sonnet-4.5 \citeyearpar{anthropic2025claude45}      & 93.31 & 87.40 & 8.80 & 8.97 & 8.93 & 8.73 & 8.37 & 9.10 & 8.97 & 8.93 & 8.13 & 8.47 \\
    Gemini-2.5-Flash-Thinking \citeyearpar{google2025gemini25} & 92.74 & 86.87 & 8.67 & 9.27 & 8.70 & 9.00 & 7.80 & 8.93 & 8.90 & 9.00 & 8.03 & 8.57 \\
    Claude-Sonnet-4-Thinking \citeyearpar{anthropic2024claude4} & 91.03 & 85.27 & 8.57 & 9.00 & 8.63 & 8.73 & 7.57 & 9.10 & 8.93 & 8.70 & 7.97 & 8.07 \\
    Claude-Sonnet-4 \citeyearpar{anthropic2024claude4}         & 91.00 & 85.24 & 8.57 & 8.80 & 8.50 & 8.70 & 7.80 & 9.03 & 8.80 & 8.80 & 8.17 & 8.07 \\
    GPT-4o-search \citeyearpar{OpenAI2023GPT4}                & 89.40 & 83.74 & 8.27 & 8.77 & 8.43 & 8.67 & 7.77 & 8.80 & 8.20 & 8.73 & 8.27 & 7.83 \\
    GPT-4o \citeyearpar{OpenAI2023GPT4}                       & 88.09 & 82.51 & 7.90 & 8.67 & 8.30 & 8.33 & 7.17 & 8.97 & 8.57 & 8.67 & 7.63 & 8.30 \\
    Gemini-1.5-Pro \citeyearpar{Reid2024Gemini15}              & 85.91 & 80.47 & 8.13 & 8.45 & 8.30 & 8.37 & 7.04 & 8.17 & 8.43 & 8.50 & 7.48 & 7.60 \\
    o3-mini \citeyearpar{openai2025o3mini}                     & 84.13 & 78.80 & 7.97 & 8.60 & 8.30 & 8.20 & 6.73 & 8.57 & 8.53 & 7.17 & 7.03 & 7.70 \\
    Claude-3.5-Sonnet \citeyearpar{Anthropic2024Claude35}      & 83.38 & 78.10 & 7.97 & 8.53 & 8.27 & 7.93 & 7.03 & 8.50 & 8.00 & 7.57 & 6.70 & 7.60 \\
    o1-mini \citeyearpar{openai2024o1}                         & 78.93 & 73.93 & 7.27 & 8.43 & 7.90 & 7.53 & 6.27 & 8.27 & 8.17 & 6.43 & 6.63 & 7.03 \\
    GPT-4-Turbo \citeyearpar{OpenAI2023GPT4}                  & 78.57 & 73.60 & 6.97 & 8.17 & 8.33 & 7.80 & 6.00 & 7.57 & 8.13 & 7.00 & 6.43 & 7.20 \\
    GPT-4-Preview \citeyearpar{OpenAI2023GPT4}                & 76.44 & 71.60 & 6.90 & 7.40 & 8.03 & 7.30 & 6.00 & 7.47 & 7.63 & 6.87 & 6.33 & 7.67 \\
    Baidu-4.0 \citeyearpar{Baidu2023Ernie40}                   & 75.08 & 70.33 & 7.27 & 7.23 & 7.67 & 7.43 & 5.63 & 6.47 & 6.80 & 7.63 & 7.80 & 6.40 \\
    Baidu-3.5 \citeyearpar{Baidu2023Ernie35}                   & 69.10 & 64.73 & 6.20 & 6.70 & 7.80 & 6.83 & 5.20 & 5.50 & 6.00 & 7.23 & 6.57 & 6.70 \\
    ChatGLM-Pro \citeyearpar{ChatGLM}                          & 69.10 & 64.73 & 5.90 & 7.07 & 7.03 & 7.90 & 5.43 & 6.33 & 5.00 & 6.67 & 5.97 & 7.43 \\
    GPT-4-Legacy \citeyearpar{OpenAI2023GPT4}                 & 66.15 & 61.96 & 6.50 & 6.73 & 6.60 & 6.73 & 5.43 & 6.10 & 6.47 & 5.30 & 5.20 & 6.90 \\
    Spark-3.0 \citeyearpar{iFlytek2023Spark30}                 & 65.62 & 61.47 & 5.77 & 6.50 & 7.27 & 7.30 & 5.70 & 5.90 & 5.03 & 6.50 & 5.23 & 6.27 \\
    Claude-3-Haiku \citeyearpar{Anthropic2024Claude3}          & 62.93 & 58.95 & 5.80 & 6.60 & 6.97 & 6.63 & 4.83 & 5.93 & 6.33 & 4.80 & 5.23 & 5.83 \\
    Gemini-Pro \citeyearpar{GeminiTeam2023Gemini}              & 58.18 & 54.50 & 4.87 & 5.43 & 7.07 & 6.43 & 5.10 & 4.50 & 4.65 & 6.33 & 4.42 & 5.70 \\
    GPT-3.5-turbo \citeyearpar{openai2023gpt35turbo}           & 55.42 & 51.91 & 4.97 & 5.37 & 6.40 & 6.47 & 4.43 & 4.67 & 5.43 & 4.20 & 4.37 & 5.60 \\
    MiniMax-ABAB5 \citeyearpar{MiniMax2024Abab65}              & 55.33 & 51.83 & 3.87 & 5.63 & 6.87 & 6.97 & 4.33 & 4.40 & 2.93 & 6.13 & 4.27 & 6.43 \\
    \hline
  \end{tabular}
  }
  \caption{Overall and Subject-Level Scores. $R_{\text{SOTA}}^{\text{model}}$ represents the relative score (0-100 scale) as defined in Equation (2), with Doubao-1.5-Thinking-Pro as the reference SOTA model. $S_{\text{model}}$ represents the absolute score (0-100 scale) as defined in Equation (1). Subject-level scores use a 10-point scale.}
\label{tab:overall-leaderboard}
\end{table*}

\section{Details in Experiment}
\label{sec:detailsInExperiment}

This section provides additional experimental details demonstrating the robustness and stability of our ranking system. We present validation experiments across different sample sizes and show the consistency of our relative scoring methodology.

\subsection{Sampling Validation}

To verify the stability of our ranking system, we conducted
evaluations across multiple sample sizes of n=1000 (in three
rounds), 2000, and 4000 questions. The results in Table~\ref{tab:test-sizes} demonstrate remarkable consistency in ranking order across all sample sizes.

Our relative scoring methodology produces smaller variance compared with absolute scoring approaches. The variance analysis reveals that top tier models show exceptional stability. DeepSeek-V3 exhibits a variance of 0.51 and O3-mini exhibits a variance of 0.95 while GPT-4o exhibits the highest variance of 1.63. Even the maximum variance represents less than two percent fluctuation indicating robust measurement precision.

The three independent one-thousand-sample runs demonstrate high reproducibility with models maintaining consistent relative positions across all test conditions. These findings validate that our ranking methodology captures stable model capabilities rather than random fluctuations.

\subsection{LLM-as-Judge Validation}

We calculated Cohen’s $\kappa$ coefficients between human evaluations and three LLM judges Doubao Gemini and GPT-4o across two evaluation rounds for thirteen models. As shown in Table~\ref{tab:model_performance_categorized_en} GPT-4o demonstrates superior performance with $\kappa$ values consistently above 0.90 with an average of 0.901 in Round 1 and 0.892 in Round 2 indicating almost perfect agreement.

The data reveal notable stability differences among judges. GPT-4o maintains consistently high agreement across both rounds with minimal variation while Doubao and Gemini exhibit more fluctuation between rounds. Specifically Doubao’s performance ranges from 0.232 to 0.745 across different models and Gemini exhibits even greater instability with some models showing dramatic drops between rounds—for example DeepSeek-V3 declines from 0.627 to 0.147.

In contrast Doubao and Gemini show lower overall agreement with average $\kappa$ values of 0.493 and 0.446 for Round 1 and Round 2 for Doubao and 0.494 and 0.400 for Gemini. Based on GPT-4o’s consistently high correlation with human evaluations and superior stability we selected it as our primary judge for reliable assessment.

\section{Implementation Details}

This section provides detailed information about the annotation processes and evaluation procedures underlying our LLMEval-Fair platform. We describe the expert involvement in data curation, validation processes, and the associated costs to ensure transparency and reproducibility.

\subsection{Data Annotation Process}

A total of 38 experts were engaged in data annotation and cleaning processes, with an average of more than 3 relevant specialists assigned to each discipline. For the annotation of original data, to mitigate fatigue-induced errors, annotation tasks for each expert were distributed across a 30–60 day period.

The cumulative remuneration disbursed to experts involved in data annotation and cleaning amounted to \$48,700. Ongoing investments are being allocated to further hire experts to expand the dataset.

\subsection{Manual Evaluation Process}

To validate our LLM-as-Judge approach, we conducted comprehensive human evaluation studies. A total of 18 experts participated in manual evaluation processes, with an average of about 2 relevant specialists assigned to each discipline. This expert-based validation ensures that our automated evaluation system maintains high agreement with human judgment standards.

The evaluation process involved multiple rounds of assessment across 13 representative models, with experts providing independent judgments that were subsequently compared against our LLM-based evaluation system using Cohen's $\kappa$ coefficient.

\subsection{Cost Analysis}

The development and validation of LLMEval-Fair required substantial investment in both human expertise and computational resources. We spent more than \$5,000 on using latest APIs of LLMs and deploying models for evaluation purposes. Additionally, \$10,000 was allocated for hiring qualified volunteers to conduct manual evaluations, ensuring rigorous validation of our automated assessment framework.

\begin{table*}[t]
    \centering
    \scalebox{0.9}{
    \begin{tabular}{lccccccccc}
    \toprule
    & \multicolumn{3}{c}{\textbf{1000 Questions}} & \multicolumn{2}{c}{\textbf{Larger Samples}} & \multicolumn{2}{c}{\textbf{Statistics}} \\
    \cmidrule(lr){2-4} \cmidrule(lr){5-6} \cmidrule(lr){7-8}
    \textbf{Model} & \textbf{Trial 1} & \textbf{Trial 2} & \textbf{Trial 3} & \textbf{2000} & \textbf{4000} & \textbf{Mean} & \textbf{Variance} \\
    \midrule
    Doubao-1.5-Thinking-Pro & 100.0 & 100.0 & 100.0 & 100.0 & 100.0 & 100.0 & 0.00 \\
    DeepSeek-V3             & 96.48 & 96.87 & 98.10 & 98.02 & 97.53 & 97.40 & 0.51 \\
    Qwen-3-32B              & 92.21 & 92.58 & 93.93 & 94.15 & 93.45 & 93.26 & 0.71 \\
    GPT-4o                  & 88.08 & 90.21 & 91.50 & 90.69 & 90.48 & 90.19 & 1.63 \\
    o3-mini                 & 84.13 & 85.31 & 86.69 & 85.56 & 86.18 & 85.57 & 0.95 \\
    \bottomrule
    \end{tabular}
    }
    \caption{Ranking stability across different sample sizes. All scores are relative scores with Doubao-1.5-Thinking-Pro as reference (100.0). The three 1000-question trials demonstrate high reproducibility, with low variance indicating robust measurement precision.}
    \label{tab:test-sizes}
\end{table*}

\begin{table*}[t]
    \centering
    \scalebox{0.85}{
    \begin{tabular}{lcccccc}
    \toprule
    & \multicolumn{3}{c}{\textbf{Round 1}} & \multicolumn{3}{c}{\textbf{Round 2}} \\
    \cmidrule(lr){2-4} \cmidrule(lr){5-7}
    \textbf{Model Name} & \textbf{Doubao} & \textbf{Gemini} & \textbf{GPT-4o} & \textbf{Doubao} & \textbf{Gemini} & \textbf{GPT-4o} \\
    \midrule
    \textit{\textbf{Open-source LLMs}} \\
    DeepSeek-R1 & 0.527 & 0.662 & 0.960 & 0.451 & 0.251 & 0.977 \\
    DeepSeek-V3 & 0.326 & 0.627 & 0.949 & 0.366 & 0.147 & 0.800 \\
    Qwen-3-235B & 0.486 & 0.468 & 0.818 & 0.232 & 0.496 & 0.931 \\
    Qwen-3-32B & 0.560 & 0.390 & 0.886 & 0.414 & 0.271 & 0.872 \\
    \midrule
    \textit{\textbf{Closed-source LLMs}} \\
    Claude-Sonnet-4 & 0.309 & 0.186 & 0.826 & 0.402 & 0.677 & 0.909 \\
    Claude-Sonnet-4-Thinking & 0.451 & 0.399 & 0.830 & 0.443 & 0.373 & 0.684 \\
    Doubao-1.5-Pro & 0.629 & 0.388 & 0.950 & 0.383 & 0.451 & 0.915 \\
    Doubao-1.5-Thinking-Pro & 0.707 & 0.786 & 0.993 & 0.745 & 0.324 & 0.920 \\
    Gemini-2.5-Pro & 0.451 & 0.539 & 0.831 & 0.376 & 0.682 & 0.858 \\
    Gemini-2.5-Pro-Thinking & 0.408 & 0.547 & 0.843 & 0.344 & 0.185 & 0.859 \\
    GPT-4o & 0.447 & 0.507 & 0.925 & 0.553 & 0.417 & 0.962 \\
    o1 & 0.599 & 0.535 & 0.933 & 0.571 & 0.555 & 0.957 \\
    o3-mini & 0.517 & 0.397 & 0.975 & 0.531 & 0.381 & 0.963 \\
    \midrule
    \textbf{Mean} & \textbf{0.493} & \textbf{0.495} & \textbf{0.902} & \textbf{0.447} & \textbf{0.401} & \textbf{0.893} \\
    \bottomrule
    \end{tabular}
    }
    \caption{Cohen's $\kappa$ correlation coefficient between human evaluation and three large language model evaluations across two rounds.}
    \label{tab:model_performance_categorized_en}
\end{table*}

\section{JWT Authentication Process}

This section describes the detailed implementation of our JSON Web Token (JWT) authentication system, which forms the outer layer of our two-tier anti-cheating architecture. The JWT process ensures secure and authenticated access to our evaluation platform while preventing unauthorized access and session manipulation.

Our JWT implementation follows a standard three-phase protocol: token generation, transmission, and verification. Algorithm~\ref{alg:jwt} outlines the complete JWT authentication workflow used in LLMEval-Fair.

\begin{algorithm}[t!]
    \caption{JWT Authentication Process in LLMEval-Fair}
    \label{alg:jwt}
    \begin{algorithmic}[1]
    \STATE \textbf{Server-side Token Generation:}
    \STATE Generate a unique user identity (\texttt{user\_id})
    \STATE Generate current timestamp and expiration time (\texttt{exp})
    \STATE Construct payload $\leftarrow$ \{\texttt{user\_id}, \texttt{timestamp}, \texttt{exp}, \texttt{session\_id}, \texttt{permissions}\}
    \STATE Sign payload with server Secret using HMAC-SHA256 to generate JWT
    \STATE \textbf{return} JWT to the authenticated user
    \STATE
    \STATE \textbf{Client-side Request:}
    \STATE Include JWT in Authorization header: \texttt{Bearer <token>}
    \STATE Send request to evaluation endpoint
    \STATE
    \STATE \textbf{Server-side Verification:}
    \STATE Extract JWT from Authorization header
    \STATE Verify JWT signature using server Secret
    \STATE Parse payload and extract claims
    \IF{JWT signature is invalid}
        \STATE \textbf{return} HTTP 401 Unauthorized
    \ENDIF
    \IF{current\_time $>$ \texttt{exp}}
        \STATE \textbf{return} HTTP 401 Token Expired
    \ENDIF
    \IF{\texttt{user\_id} not found \textbf{or} permissions insufficient}
        \STATE \textbf{return} HTTP 403 Forbidden
    \ENDIF
    \IF{session validation fails (e.g., concurrent sessions detected)}
        \STATE \textbf{return} HTTP 403 Session Invalid
    \ENDIF
    \STATE Allow evaluation operation to proceed
    \STATE Log access attempt with \texttt{user\_id}, \texttt{timestamp}, and \texttt{session\_id}
    \end{algorithmic}
\end{algorithm}

\section{Non-Monotonic Performance Trends}
\label{sec:non_monotonic}

Performance trajectories within a model family are not always monotonically increasing across versions. For example, DeepSeek-V3.2 scores lower than DeepSeek-V3 despite being released later. We identify three factors that explain such non-monotonic trends:

\textbf{Training reward changes.} DeepSeek-V3.2 removed format rewards and introduced generative reward models compared to V3, leading to outputs that favor comprehensive discussion over the precision required by exam-style questions. In short-answer questions, V3 produces concise, targeted responses while V3.2 tends to embed correct answers within verbose explanations, resulting in lower scores under our scoring rubric.

\textbf{Subject-specific training data bias.} The distribution of disciplinary data varies across model training corpora. Later versions may improve on some subjects while regressing on others due to shifts in training data composition.

\textbf{Model specification trade-offs.} Different versions within the same family (e.g., lightweight vs.\ full, reasoning-optimized vs.\ general-purpose) involve trade-offs in generation quality. Performance should be compared at equivalent scale and optimization objectives rather than purely by release chronology.

These observations underscore the value of domain-specific dynamic benchmarks for revealing capability trade-offs that aggregate leaderboard scores may obscure.

\section{Augmentation Validation}
\label{sec:augmentation_validation}

To verify that our question augmentation process preserves evaluation validity, we conduct a controlled comparison between original and augmented questions. We trace augmented questions back to their originals and construct paired subsets covering the same knowledge points. Five representative models spanning different capability tiers are evaluated on both subsets under identical conditions, with results reported in Table~\ref{tab:aug_validation}.

\begin{table}[ht]
  \centering
  \scalebox{0.85}{
  \begin{tabular}{lcc c}
    \toprule
    \textbf{Model} & \textbf{Original} & \textbf{Augmented} & $\Delta$ \\
    \midrule
    Gemini-2.5-Pro         & 92.59 & 91.85 & $-$0.74 \\
    DeepSeek-V3            & 92.48 & 91.04 & $-$1.44 \\
    Claude-Sonnet-4        & 89.65 & 86.07 & $-$3.59 \\
    GPT-4o                 & 84.62 & 81.40 & $-$3.22 \\
    o3-Mini                & 74.86 & 80.75 & $+$5.89 \\
    \bottomrule
  \end{tabular}
  }
  \caption{Performance comparison between original and augmented questions.}
  \label{tab:aug_validation}
\end{table}

Statistical analysis confirms no significant difference: the paired $t$-test yields $p = 0.736$ with a 95\% confidence interval of $[-5.37, +4.13]$, and Cohen's $d = 0.16$ (negligible effect). The model ranking under original questions is identical to that under augmented questions (Spearman $\rho = 1.0$). Four of five models score slightly lower on augmented questions, consistent with the removal of answer-choice scaffolding that slightly increases difficulty. The exception is o3-Mini ($+$5.89), which benefits from the fill-in-the-blank format that eliminates distractors. Top-tier models show the smallest sensitivity (Gemini: $-$0.74, DeepSeek: $-$1.44), suggesting they rely on genuine knowledge rather than format-specific cues.

\section{Error Categorization Methodology}
\label{sec:error_categorization}

Our five error categories were developed through a systematic qualitative coding process. First, we sampled error responses across multiple models and conducted open coding to identify recurring failure patterns (\textbf{pilot coding}). Through iterative refinement, we consolidated the patterns into five categories (\textbf{codebook development}): (1)~disciplinary knowledge gaps, (2)~misunderstanding, (3)~logical reasoning errors, (4)~factual inaccuracies, and (5)~format compliance failures, each with an operational definition and decision rules. During \textbf{annotation}, each erroneous response was assigned exactly one primary label. When multiple failure modes co-occurred, a fixed priority ordering was applied (disciplinary knowledge $>$ misunderstanding $>$ logical reasoning $>$ factual inaccuracies $>$ format compliance) to ensure consistency.

\begin{itemize}
    \item \textbf{Disciplinary knowledge gaps}: The model lacks domain-specific knowledge required to answer the question (e.g., missing a key medical term or legal principle).
    \item \textbf{Misunderstanding}: The model misinterprets the question's intent or misses decisive contextual cues (e.g., overlooking the keyword ``first'' in a question asking for the earliest example).
    \item \textbf{Logical reasoning errors}: The model possesses the relevant knowledge but applies incorrect reasoning chains or draws invalid conclusions.
    \item \textbf{Factual inaccuracies}: The model generates statements that contradict established facts, despite understanding the question correctly.
    \item \textbf{Format compliance failures}: The model provides a substantively correct answer but fails to conform to the required response format (e.g., providing a narrative when a list is expected).
\end{itemize}

\section{Cross-Judge Experiment}
\label{sec:cross_judge}

To validate that GPT-4o introduces no systematic bias as judge, three judges from different model families---GPT-4o (OpenAI), Doubao-1.5-Thinking-Pro (ByteDance), and Gemini-2.5-Pro (Google)---independently scored the same 1,300 samples (100 questions $\times$ 13 models) on the 0--3 scale. Table~\ref{tab:cross_judge_ranking} reports inter-judge ranking consistency, Table~\ref{tab:cross_judge_item} reports item-level agreement, and Table~\ref{tab:family_bias} reports the family bias analysis.

\begin{table}[ht]
  \centering
  \scalebox{0.85}{
  \begin{tabular}{lcccc}
    \toprule
    \textbf{Judge Pair} & $\rho$ & $p$ & $\tau$ & $p$ \\
    \midrule
    GPT-4o vs.\ Doubao  & 0.894 & $<$0.001 & 0.763 & $<$0.001 \\
    GPT-4o vs.\ Gemini  & 0.847 & $<$0.001 & 0.693 & 0.001 \\
    Doubao vs.\ Gemini  & 0.900 & $<$0.001 & 0.782 & $<$0.001 \\
    \bottomrule
  \end{tabular}
  }
  \caption{Inter-judge ranking consistency (Spearman $\rho$ and Kendall $\tau$).}
  \label{tab:cross_judge_ranking}
\end{table}

\begin{table}[ht]
  \centering
  \scalebox{0.85}{
  \begin{tabular}{lccc}
    \toprule
    \textbf{Judge Pair} & \textbf{Exact} & $\pm$\textbf{1 Tol.} & \textbf{Wt.\ $\kappa$} \\
    \midrule
    GPT-4o vs.\ Doubao  & 78.4\% & 96.9\% & 0.649 \\
    GPT-4o vs.\ Gemini  & 81.5\% & 94.6\% & 0.603 \\
    Doubao vs.\ Gemini  & 74.5\% & 94.5\% & 0.543 \\
    \bottomrule
  \end{tabular}
  }
  \caption{Item-level agreement between judge pairs.}
  \label{tab:cross_judge_item}
\end{table}

\begin{table}[ht]
  \centering
  \scalebox{0.80}{
  \begin{tabular}{lccccccc}
    \toprule
    \textbf{Judge} & \textbf{Own} & \textbf{Own Avg} & \textbf{3-Judge} & $\Delta$ \\
    \midrule
    GPT-4o  & OpenAI    & 2.47 & 2.39 & $+$0.08 \\
    Doubao  & ByteDance & 2.68 & 2.75 & $-$0.07 \\
    Gemini  & Google    & 2.79 & 2.67 & $+$0.12 \\
    \bottomrule
  \end{tabular}
  }
  \caption{Family bias test: each judge's average score for its own family vs.\ the three-judge mean. All differences are negligible ($|\Delta| \leq 0.12$ on a 0--3 scale).}
  \label{tab:family_bias}
\end{table}

\section{Evaluation Timeline}
\label{sec:eval_timeline}

Each model entry corresponds to a specific, fixed API snapshot. Table~\ref{tab:timeline_openai} provides the OpenAI family timeline as an example; equivalent tables for other families are available in our released materials.

\begin{table}[ht]
  \centering
  \scalebox{0.75}{
  \begin{tabular}{llcc}
    \toprule
    \textbf{Model} & \textbf{API Snapshot} & $S_{\text{model}}$ & \textbf{Eval.} \\
    \midrule
    GPT-3.5-Turbo     & gpt-3.5-turbo-0613      & 51.90 & 2023 H1 \\
    GPT-4              & gpt-4-0613              & 61.97 & 2023 H1 \\
    GPT-4-Turbo        & gpt-4-1106-preview      & 73.60 & 2023 H2 \\
    GPT-4-Preview      & gpt-4-0125-preview      & 71.60 & 2024 H1 \\
    o1-Mini            & o1-mini-2024-09-12      & 73.93 & 2024 H2 \\
    GPT-4o             & gpt-4o-2024-11-20       & 82.50 & 2024 H2 \\
    GPT-4o-Search      & gpt-4o-search-2024-11-20 & 83.73 & 2024 H2 \\
    o1                 & o1-2024-12-17           & 87.43 & 2024 H2 \\
    o3-Mini            & o3-mini-2025-01-31      & 78.80 & 2025 H1 \\
    GPT-5              & gpt-5-2025-08-07        & 87.90 & 2025 H1 \\
    \bottomrule
  \end{tabular}
  }
  \caption{Evaluation timeline for the OpenAI model family. $S_{\text{model}}$ is the absolute score. Eval.\ denotes the evaluation window.}
  \label{tab:timeline_openai}
\end{table}

\end{document}